\def\year{2020}\relax
\documentclass[letterpaper]{article} 
\usepackage{aaai}  
\usepackage{times}  
\usepackage{helvet} 
\usepackage{courier}  
\usepackage[hyphens]{url}  
\usepackage{graphicx} 
\usepackage{graphicx}  
\usepackage{multirow}
\usepackage{amsmath}
\usepackage{amssymb}
\usepackage{algorithm}
\usepackage{verbatim}
\usepackage{algpseudocode}
\usepackage{graphicx}
\usepackage{subfigure}
\usepackage{booktabs}
\usepackage{xcolor}
\newtheorem{thm}{Theorem}[section]
\newtheorem{lem}[thm]{Lemma}
\newtheorem{prop}[thm]{Proposition}

\newtheorem{defn}{Definition}[section]

\usepackage{hyperref}

\title{Robust Generative Adversarial Network}

\author{Shufei Zhang\\
University of Liverpool\\
Shufei.Zhang@xjtlu.edu.cn\\
\And Zhuang Qian\\
University of Liverpool\\
Kaizhu.Huang@xjtlu.edu.cn\\
\And Kaizhu Huang\\
University of Liverpool\\
Kaizhu.Huang@xjtlu.edu.cn
\AND Jimin Xiao\\
University of Liverpool\\
Jimin.Xiao@xjtlu.edu.cn \\
\And Yuan He\\
Alibaba Group\\
heyuan.hy@alibaba-inc.com
}

\begin{document}

\maketitle
\begin{abstract}
   Generative adversarial networks (GANs) are powerful generative models, but usually suffer from instability and generalization problem which may lead to poor generations. Most existing works 
  focus on stabilizing the training of the discriminator 
  while ignoring the generalization properties. 
  In this work, we aim to improve the generalization capability of GANs by promoting the local robustness within the small neighborhood of the training samples. We also prove that the robustness in small neighborhood of training sets can lead to better generalization.
  Particularly, we design a robust optimization framework where
  the generator and discriminator compete with each other in a \textit{worst-case} setting within a small Wasserstein ball.  The generator tries to map  \textit{the worst input distribution} (rather than a Gaussian distribution used in most GANs) to the real data distribution, while the discriminator attempts to distinguish the real and fake distribution \textit{with the worst perturbation}.
  We have proved that our robust method can obtain a tighter generalization upper bound than traditional GANs under mild assumptions, ensuring a theoretical superiority of RGAN over GANs.
  A series of experiments on CIFAR-10, STL-10 and CelebA datasets indicate that our proposed robust framework can improve on five baseline GAN models substantially and consistently.
\end{abstract}

\section{Introduction}

Generative adversarial networks (GANs) \cite{goodfellow2014generative} have been enjoying much attention recently due to their great success on different tasks and datasets \cite{radford2015unsupervised,salimans2016improved,ho2016generative,li2017adversarial}
The framework of GANs can be formulated as the game between generator and discriminator. The generator tries to produce the fake distribution which approximates the real data distribution, while the discriminator attempts to distinguish the fake distribution from the real distribution. These two players compete with each other iteratively. GANs are also popular for their theoretical value \cite{nowozin2016f,uehara2016generative,mohamed2016learning}.


Despite the success, traditional GANs usually suffer from training instability and low generalization capability which leads to poor generation performance. Most researches focus on how to stabilize the training of GANs \cite{arjovsky2017wasserstein,miyato2018spectral,gulrajani2017improved}.
However,
the generalization property of GANs is not well understood.  
Several works showed that if the discriminator has poor generalization, the generator cannot learn effectively \cite{goodfellow2014generative,gan_principle} and are prevented from learning the target distribution which leads to poor generations \cite{improving_generalization}. To tackle such problem, recent work \cite{improving_generalization} proposed gradient penalty to regularizing the discriminator for improving generalization. Differently, we attempt to analyze the generalization capability of GAN from new prospective of local robustness and improve the generalization by promoting the robustness within small neighborhood of training data. Moreover, we also provide the corresponding upper bound for generalization.

The generalization of discriminator/generator is to describe how well it performs on unseen data \cite{gen_equ,robust_gen,improving_generalization}. Therefore, it can be measured by the difference between its performance on training data and the whole data space. Specifically, \cite{qi2017loss,robust_gen} directly evaluate it by the difference between loss functions on training set and the whole data set. In this paper, we use the same way to evaluate the generalization for both discriminator and generator as \cite{qi2017loss,robust_gen}.

Specifically, we propose a novel method called robust generative adversarial network (RGAN) where the generator and discriminator still compete with each other iteratively, but \textit{in a worst-case setting} with respect to all the possible distributions within a small region. Specifically, a robust optimization is designed with considering the worst distribution within a small Wasserstein ball. The generator tries to map the \textit{worst input distribution} with respect to all the possible distributions within a small region (rather than a specific distribution as used in traditional GANs) to the real data distribution, while the discriminator attempts to distinguish the real and fake distribution \textit{with the worst perturbation}. Under very mild assumptions, we prove that the proposed Robust GAN attains a tighter generalization upper bound than traditional GANs, making our model theoretically more attractive. We compare our robust framework with various popular GANs (e.g., DCGAN, WGAN-GP, SNGAN, and BWGAN) \cite{radford2015unsupervised,adler2018banach,miyato2018spectral}, all of which intend to control the discriminator only. We observe substantial improvements over these models consistently on all the datasets used in this paper.\footnote{Codes of RGAN can be downloaded in the supplementary material of the submission, which will also be publicly available after the review process.}


In a nutshell, our contributions are as follows:
1)	We proposed a novel robust method called robust generative adversarial network (RGAN) which ensures the better generalization than original GANs with theoretical gaurantees.
2)  We analyse the relationship between robustness and generalization and provide corresponding upper bound. We show that robustness in sufficient small neighborhood of training data can boost generalization.
3)	The optimization procedure is simple (just inject specific perturbations) but effective.
4)	Our method can be easily implemented on different frameworks and achieves consistent improvement.



\section{Related Work}
Arjovsky et al. proved that the supports of the fake and real distributions are typically disjoint on low dimensional manifolds and there is a nearly trivial discriminator which can correctly classify the real and fake data \cite{arjovsky2017wasserstein}. The loss of such discriminator converges quickly to zero which causes the vanishing gradient for generator, further leading to the instability problem. To alleviate such problem, clipping weight and gradient penalty based methods are proposed to stabilizing the discriminator.
\cite{gulrajani2017improved,miyato2018spectral,arjovsky2017wasserstein,gan_stable,gan_convergence,gan_regularization}. Different from these works, our method aims at improving generalization of GANs.

Several works have analyzed the generalization of GANs. \cite{gen_equ} provided the upper bound for generalization of discriminator with neural network distance and prove that such distance ensure the better generalization than Jensen-Shannon divergence and Wasserstein distance. \cite{qi2017loss} analyze the upper bound of generalization for both generator and discriminator of Loss-Sensitive GAN. \cite{gan_mode} proposed to measure the generalization capability of GAN by estimating the number of modes in the model distribution. Recently, \cite{improving_generalization} attempted to improve the generalization with gradient penalty. However, those previous works do not consider the relationship between local robustness and generalization. Our method improve the generalization from perspective of robustness and show superior performance than methods based on gradient penalty.

There are also some works analyzing the relationship between robustness and generalization. \cite{Disentangling} showed that on-manifold robustness is essentially generalization with experiments. However, the theoretical bound is not provided. \cite{robust_gen} state the relationship between robustness and generalization theoretically. The defect is that such theory is hard to apply directly. Compared with previous works, we provide the stronger theoretical relationship between robustness and generalization. We also provide the upper bound for generalization which can be directly used to compare the generalization capability of two models.

The Robust Conditional Generative Adversarial Networks has been recently proposed to improve the robustness of conditional GAN for noised data. However, this method can merely implement on conditional GAN which improves the ability of generator only to defend the noise \cite{chrysos2018robust}. The generalization is not considered. Some other researchers also focus on the robustness of GAN to label noise \cite{thekumparampil2018robustness }\cite{kaneko2019label}. In comparison, we propose a comprehensive study and focus on the model robustness of both discriminator and generator to input noise with a guaranteed tighter bound than traditional GANs.

\section{Generative Adversarial Network}

Formally, the training procedure of GAN can be formulated as:
\begin{eqnarray}
\begin{aligned}
\min_{G}\max_{D}S(G,&D) \triangleq \mathbb{E} _{x\sim P_r}[logD(x)]\\ &+\mathbb{E}_{\widetilde{x}\sim P_g}[(1-logD(G(z)))]
\label{eq:1}
\end{aligned}
\end{eqnarray}
where $x$ and $\widetilde{x}=G(z)$ are real and fake examples sampled from the real data distribution $P_r$ and generation distribution $P_g$ respectively. The generation distribution is defined by $G(z)$ where $z\sim P_z$ ($P_z$ is a specific input noise distribution). The minmax problem cannot be solved directly since the expectation of the real and generation distribution is usually intractable. Therefore, an approximation problem can be defined as:
\begin{eqnarray}
\begin{aligned}
\min_{G}\max_{D} S_m(G,&D) \triangleq \frac{1}{m}\sum_{i=1}^{m}[logD(x_i)]\\ &+\frac{1}{m}\sum_{i=1}^{m}[(1-logD(G(z_i)))]
\label{eq:2}
\end{aligned}
\end{eqnarray}
where $m$ examples of $x_i$ and $z_i$ are sampled from two specific distributions $P_r$ and $P_z$. The mean value of the loss is used to approximate the original problem. However, robustness is not adequately considered especially when specific distributions $P_r$ and $P_z$ are assumed. This may sometimes cause the instability issue and further lead to degraded generations.
In this paper, to alleviate such problem, we design a distributionally robust optimization. Particularly, we consider the worst distributions (rather than a specific single distribution) so as to achieve a robust GAN.

\section{Robust Generative Adversarial Network}

In this section, we design the distributionally robust optimization  and develop the novel robust framework on GANs. 

\subsection{Distributionally Robust Optimization}\label{sec:dro}

Let $d: \mathcal{X}\times \mathcal{X} \to \mathbb{R}_{+}$. We define $d(x,x_0)$ as the departure between $x$ and $x_0$ (with $x$, $x_0\in \mathcal{X}$). For distributionally robust optimization, we consider the robustness region $\mathcal{P}=\{P:W(P,P_0)\leq \rho\}$ which describes the $\rho$ -neighborhood of the distribution $P_0$ under the divergence $W(.,.)$. Normally,  $W(.,.)$  can be defined as the Wasserstein metric where the corresponding $d(x,x_0)=\|x-x_0\|_2^2$.\footnote{Though we focus on using 2-norm in this paper, other norms like 1-norm or $\infty$-norm can also be used.}  In comparison from the single distribution $P_0$ (typically used in traditional GANs), the robustness region $\mathcal{P}=\{P:W(P,P_0)\leq \rho\}$ can usually lead to better robustness.

The distributionally robust optimization can be formulated as~\cite{sinha2017certifying}:
\begin{eqnarray}
\begin{aligned}
\min_{\theta} \sup_{P\in \mathcal{P}} \mathbb{E}_{P}[l(\mathcal{X};\theta)]
\label{eq:ori}
\end{aligned}
\end{eqnarray}
where $l(.)$ is a loss function parameterized  by $\theta$. The problem of (\ref{eq:ori}) is typically intractable for arbitrary $\rho$.  We can reformulate it into its Langrangian relaxation as follows:
\begin{equation}
\min_{\theta}\sup_{P\in \mathcal{P}}\{\mathbb{E}_P[l(\mathcal{X};\theta)]-\lambda W(P,P_0)\}\label{eq:lag}
\end{equation}
where $\lambda$ is the Langrangian constant parameter.

In order to solve this problem, we present a proposition:
\begin{prop}
Let $l$: $\mathcal{\theta}\times \mathcal{X}\to \mathbb{R}$ and $d$: $\mathcal{X}\times \mathcal{X}\to \mathbb{R}_+$ be continuous. Then, for any distribution $P_0$ and $\rho>0$ we have
\begin{eqnarray}
\begin{aligned}
\sup_{P\in \mathcal{P}}\{\mathbb{E}_P[l(\mathcal{X};&\theta)]-\lambda W(P,P_0)\}\\ &=\mathbb{E}_{P_0}[\sup_{x\in \mathcal{X}}\{l(x;\theta)-\lambda d(x,x_0)\}]
\label{eq:7}
\end{aligned}
\end{eqnarray}

(Proof is provided in~\cite{sinha2017certifying}).
\label{lem_l}
\end{prop}

With Proposition~\ref{lem_l}, we can transform (\ref{eq:lag})  as follows:
\begin{eqnarray}
\begin{aligned}
\min_{\theta} \mathbb{E}_{P_0}\sup_{x\in \mathcal{X}}[l(x;\theta)-\lambda d(x,x_0)]
\label{eq:3}
\end{aligned}
\end{eqnarray}
where the second term $d(x,x_0)$ is to restrict the distance between two points.

\subsection{Robust Training  over Generator}
With the distributionally robust optimization, we first discuss how we can perform robust training over generator. The generator of traditional GANs tries to map a specific noise distribution $P_z$ (typically a Gaussian distribution) to the image distribution $P_r$. The objective of generator is described as follows:
\begin{eqnarray}
\begin{aligned}
\min_{G}\frac{1}{m} \sum_{i=1}^{m}[log(1-D(G(z_i)))], \quad \mbox{where}\quad  z_i\sim P_z
\label{eq:4}
\end{aligned}
\end{eqnarray}
For improving the robustness, we consider all the possible distributions within the robustness region $\mathbb{P}_z=\{P:W(P,P_z)\leq \rho_z\}$ rather than a single specific distribution where $W$ is defined as the Wasserstein metric measuring the distance between $P$ and $P_z$.  To enable feasible optimization,  we further consider to minimize the  upper bound of (\ref{eq:4}) for all the possible distributions within the robustness region. Namely, we tend to minimize the worst  distribution within the robustness region:
\begin{eqnarray}
\begin{aligned}
\min_{G}\sup_{P\in \mathbb{P}_z}\frac{1}{m}\sum_{i=1}^{m}[log(1-D(G(z_i)))],\; \mbox{where}\; z_i\sim P
\label{eq:5o}
\end{aligned}
\end{eqnarray}

With the similar method discussed in  Section~\ref{sec:dro}, we can also relax (\ref{eq:5o}) as:
\begin{eqnarray}
\begin{aligned}
&\min_{G}\max_{r}\frac{1}{m}\sum_{i=1}^{m}[log(1-D(G(z_i+r^i)))-\lambda_z\|r^i\|_2^2]  \\ & \mbox{where}\quad z_i\sim P_z\; \mbox{and the trade-off parameter\;} \lambda_z>0.
\label{eq:5}
\end{aligned}
\end{eqnarray}
Different from those previous methods, our method attempts to map the worst distribution (in the  $\rho_z$-neighborhood of the original distribution $P_z$) to the image distribution. Intuitively, we tend to learn a good $G$ that can even perform well on the worst-case noisy points or the most risky points. Therefore, such generator would be robust against poor input noises and might be less likely to generate the low-quality images.

\subsection{Robust Training over Discriminator}
In traditional GANs described by (\ref{eq:2}), the generator attempts to generate a fake distribution to approximate the real data distribution, while the discriminator tries to learn the decision boundary to separate real and fake distributions. Apparently, a discriminator with a poor robustness would inevitably mislead the training of generator. In this section, we utilize the popular adversarial learning method~\cite{FSGM,feature_scattering} and propose the robust optimization method to improve the discriminator's robustness both for clean and noised data.

Specifically, we define the robust regions for both the fake distribution $\mathbb{P}_g=\{P:W(P,P_g)\leq \rho_g\}$ and real distribution $\mathbb{P}_r=\{P:W(P,P_r)\leq \rho_r\}$. The generator tries to reduce the distance between the fake distribution $P_g$ and real distribution $P_r$. The discriminator attempts to separate the worst distributions in $\mathbb{P}_g$ and $\mathbb{P}_r$. Intuitively, the worst distributions are closer to the decision boundary (less discriminative) and they are able to guide the training of discriminator to perform well on "confusing" data points near the classification boundary (such discriminator can be more robust than original one). We can reformulate the optimization of $D$ in~(\ref{eq:2}) as the following robust version:
\begin{eqnarray*}
\footnotesize{
\begin{aligned}
\max_{D}\sup_{P_1\in \mathbb{P}_r}\frac{1}{m}&\sum_{i=1}^{m}[logD(x'_i)]+\sup_{P_2\in \mathbb{P}_g}\frac{1}{m}\sum_{i=1}^{m}[log(1-D(G'(z_i)))]
\label{eq:7}
\end{aligned}
}
\end{eqnarray*}
 where $z_i\sim P_z$, $x'_i\sim P_1$, and $G'\sim P_2$.
Similarly, with the method discussed in Section~\ref{sec:dro}, we can relax this problem as:
\begin{eqnarray}
\footnotesize{
\begin{aligned}
\max_{D}\min_{r_{1},r_{2}}\frac{1}{m}&\sum_{i=1}^{m}[logD(x_i+r_{1}^i)]\\ &+\frac{1}{m}\sum_{i=1}^{m}[log(1-D(G(z_i)+r_{2}^i))]
\\ &+\frac{\lambda_d}{m}\sum_{i=1}^{m} [\|r^i_{1}\|_2^2]+\|r^i_{2}\|_2^2] \\ &\mbox{with} \quad z_i\sim P_z,\quad x_i\sim P_r
\label{eq:7}
\end{aligned}}
\end{eqnarray}

Here $r_{1}=\{r_{1}^i\}_{i=1}^m$ is the set of small perturbations for the points sampled from real distribution $P_d$ which try to make the real distribution closer to the fake distribution. $r_{2}=\{r_{2}^i\}_{i=1}^m$ tries to make the fake distribution closer to the real one. Intuitively, these perturbations try to enhance the difficulty of the classification task for discriminator by making the real and fake data less distinguishable. A discriminator trained with the difficult tasks would tend to be robust in practice. This has been proved theoretically in the next section.
\begin{algorithm}[tb]
\centering
\caption{Algorithm for RGAN}
\label{alg:Framwork}
\begin{algorithmic}[1]
\State {\bfseries Input:} size $m$, trade-off parameter $\lambda$, perturbation magnitude (respectively for $G$ and $D$) $\epsilon_1$, $\epsilon_2$
\For{number of training iterations}
\State Sample a batch of input noise $z_i\sim P_z$ of size $m$, a batch of real data $x_i\sim P_r$ of size $m$. $\lambda$ is the trade-off parameter for original objective and our objective. $\epsilon_1$ and $\epsilon_2$ are amplitude of perturbation for input and images respectively.
\State Find the worst perturbation $\{r_{zadv},r_{dadv1},r_{dadv2}\}$ by maximizing the objective of generator and minimizing the objective of discriminator: \\
\footnotesize{
 $r^i_{zadv}=\mathop{\arg\min}_{r^i:\|r^i\|_2=1}[log(1-D(G(z_i+r^i)))+\lambda_z\|r^i\|_2^2]$\\
 $r^i_{dadv1}=\mathop{\arg\min}_{r^i:\|r^i\|_2=1}[logD(x_i+r^i)+\lambda_d\|r^i_{1}\|_2^2]$\\
 $r^i_{dadv2}=\mathop{\arg\min}_{r^i:\|r^i\|_2=1}[log(1-D(G(z_i)+r^i))+\lambda_d\|r^i_{2}\|_2^2]$}
\State Update $G$ by descending along its stochastic gradient:\\
 \footnotesize{
 $\nabla_{\theta_g}[\frac{1}{m}\sum_{i=1}^{m}[log(1-D(G(z_i)))+\frac{\lambda}{m}\sum_{i=1}^{m}[log(1-D(G(z_i+\epsilon_1 r_{zadv}^i)))$}
\State Update $D$ by descending along its stochastic gradient:\\
\footnotesize{
$\nabla_{\theta_d}[S_m(G,D)+\frac{\lambda}{m}\sum_{i=1}^{m}[logD(x_i+\epsilon_2 r_{dadv1}^i)]+\frac{\lambda}{m}\sum_{i=1}^{m}[log(1-D(G(z_i)+\epsilon_2 r_{dadv2}^i))]
]$}
\EndFor
\end{algorithmic}
\end{algorithm}
\subsection{Overall Optimization}
We now integrate the robust training of generator and discriminator into a single framework:
 \begin{eqnarray}
 \footnotesize{
\begin{aligned}
\min_{G}\max_{D}&V(G,D)\triangleq (1-\lambda)S(G,D)\\ &+\sup_{P:W(P,P_r)\leq \rho_r}\lambda\mathbb{E} _{x\sim P}[logD(x)]\\&+\sup_{P:W(P,P_g)\leq \rho_g}\lambda\mathbb{E}_{G'\sim P}[(1-logD(G'(z)))]
\label{eq:total}
\end{aligned}}
\end{eqnarray}
where $G'(z_i)=G(z_i)+r_2^i$ and  $z_i\sim p^{\lambda}_z$. $p^{\lambda}_z$ is the mixture distribution defined by $p^{\lambda}_z=(1-\lambda)p_z+\lambda p'_z$ and $p'_z$ is the worst distribution defined by $p'_z=\arg\max_{P:W(P,P_z)\leq \rho_z}\mathbb{E} _{x\sim P}[1-logD(G(x))]$. $r_2^i$ is an arbitrary perturbation.
It is noted that we also combine the original GAN into the framework, allowing a more flexible training. The specific algorithm can be seen in Algorithm~\ref{alg:Framwork}:

\section{Theoretical Analysis}
In this section, we  provide theoretical analysis for the RGAN while leaving the proof details  in the complementary material due to the space limitation. We will show that our proposed robust method RGAN can attain a tighter generalizability upper bound than traditional GANs under reasonable mild assumptions, ensuring a theoretical superiority of RGAN over GANs. The main theorems are described in Theorem~\ref{lem_d_final} and Theorem~\ref{lem_g_final}. Before this, we first set out some preliminary lemmas.


\begin{lem}\label{lem_d1}
For an arbitrary fixed G, the optimal D of the game associated with RGAN (defined by the utility function
$V(G,D)$ in (\ref{eq:total})) is:
$D^*_G(x)=\frac{p_r^{\lambda}(x)}{p_r^{\lambda}(x)+p_g^{\lambda}(x)}$
where $p_r^{\lambda}(x)=(1-\lambda)p_r+\lambda p'_r$ is the mixture distribution for real data with $\lambda\in [0,1]$, $p'_r$ is the worst distribution defined by $p'_r=\arg\min_{P:W(P,P_r)\leq \rho_r}\mathbb{E} _{x\sim P}[logD(x)]$, and $p_g^{\lambda}(x)=(1-\lambda)p_g+\lambda p'_g$ is the mixture distribution for fake data. The worst distribution $p'_g$ is defined by $p'_g=\arg\min_{P:W(P,P_g)\leq \rho_g}\mathbb{E} _{G'\sim P}[1-logD(G'(z))]$.

\end{lem}

\begin{lem}
When the optimum discriminator $D^*$ is achieved, the utility function of RGAN defined by
$V(G,D)$ in (\ref{eq:total})) reaches the global minimum if and only if $p_g^{\lambda}(x)=p_r^{\lambda}(x)$.
\label{lem_d2}
\end{lem}

The min-max problem of (\ref{eq:total}) is computationally intractable due to the expectations over real and fake distributions. An alternate way is to approximate the original problem with the empirical average of finite examples:
\begin{eqnarray}
\footnotesize{
\begin{aligned}
&\min_{G}\max_{D}V_m(G,D)\triangleq (1-\lambda)S_m(G,D)\\ &+\frac{\lambda}{m}\sum_{i=1}^m [logD(x'_i)]+\frac{\lambda}{m}\sum_{i=1}^m[(1-logD(G'(z_i)))]
\label{eq:mean}
\end{aligned}}
\end{eqnarray}
where $x'_i\sim p'_r$, $G'\sim p'_g$ and $z_i\sim p^{\lambda}_z$. $p^{\lambda}_z$ is the mixture distribution defined by $p^{\lambda}_z=(1-\lambda)p_z+\lambda p'_z$ and $p'_z$ is the worst distribution defined by $p'_z=\arg\max_{P:W(P,P_z)\leq \rho_z}\mathbb{E} _{x\sim P}[1-logD(G(x))]$.

We now provide the analysis for the generalizability of the discriminator in  Theorem~\ref{lem_d_final}. The generalizability of the discriminator in our RGAN is  defined in a way similar to that in~\cite{qi2017loss}\cite{arora2017generalization}, which describe how fast the difference $|V^\theta_m-V^\theta|$ converges, where $V^\theta=\max_{D}V(G^*,D)$ and $V^\theta_m=\max_{D}V_m(G^*,D)$. Similarly, the generalizability of the discriminator in the original GAN is defined by $|W^\theta_m-W^\theta|$, where $W^\theta=\max_{D}S(G^*,D)$ and $W^\theta_m=\max_{D}S_m(G^*,D)$. First, we give some definitions:
\begin{defn}
Algorithm $\mathcal{A}$ is $(K,\epsilon(S))$-robust, if data manifold $\mathcal{Z}$ can be partitioned into $K$ disjoint sets, denoted as $\{C_i\}^K_{i=1}$, such that $\forall$ $s\in S$, 
\begin{eqnarray}
\footnotesize{
\begin{aligned}
s,z\in C_i \longrightarrow |l(\mathcal{A}_s,s)-l(\mathcal{A}_s,z)|\leq \epsilon(S)
\label{eq:7}
\end{aligned}}
\end{eqnarray}
where $l(.)$ is the loss function and $S$ is the training set.

\label{assum}
\end{defn}

\begin{defn}
Algorithm $\mathcal{A}$ is $(\epsilon,S,\sigma)$ adversarial robust, if within the $\epsilon$-neighborhood of training set $S$, 
we have
\begin{eqnarray}
\footnotesize{
\begin{aligned}
\max_{s_a,s_b\in S, s'_b\in B(s_b,\epsilon)}|l(\mathcal{A},s_a,y_{s_a})-l(\mathcal{A},s_b,y_{s_b})|\leq \sigma
\label{eq:7}
\end{aligned}}
\end{eqnarray}
where $y_{s_b}$ and $y_{s_a}$ are labels of training samples $s_a$ and $s_b$. $s'_b$ is a perturbed sample in the region $B(s_b,\epsilon)$.
\label{assum}
\end{defn}

\begin{thm}
If the training set $S_d$ for discriminator consists of $n$ i.i.d samples, the discriminator of RGAN $D_r$ and the original GAN $D_{org}$ are both $(K,\epsilon(S_d))$ robust, $D_r$ is $(\epsilon,S_d,\sigma_{r})$ adversarial robust, and $D_{org}$ is $(\epsilon,S_d,\sigma_{org})$ adversarial robust, then, for any $\delta>0$ and  small enough $\epsilon$, with the probability at least $1-\delta$,  we have
\begin{eqnarray}
\footnotesize{
\begin{aligned}
&|V^\theta_m-V^\theta|\leq \gamma_1 \sigma_r + (1-\gamma_1)\epsilon(S_d) + M\sqrt{\frac{2Kln2+2ln\frac{1}{\delta}}{n}}
\nonumber\\
&|W^\theta_m-W^\theta|\leq \gamma_2 \sigma_{org} + (1-\gamma_2)\epsilon(S_d) \\ &+ M\sqrt{\frac{2Kln2+2ln(1/\delta)}{n}}
\nonumber
\label{eq:72}
\end{aligned}
}
\end{eqnarray}
$D_r$ obtains the tighter upper bound, i.e.:
\begin{eqnarray}
\footnotesize{
\begin{aligned}
\gamma_1& \sigma_r + (1-\gamma_1)\epsilon(S_d) + M\sqrt{\frac{2Kln2+2ln(1/\delta)}{n}} \\ &\leq \gamma_2 \sigma_{org} + (1-\gamma_2)\epsilon(S_d) + M\sqrt{\frac{2Kln2+2ln(1/\delta)}{n}}
\nonumber
\label{eq:73}
\end{aligned}}
\end{eqnarray}
where $\gamma_1,\gamma_2\in [0,1]$, which are closely related to $\sigma_r, \sigma_{org}$ and the intersection of $\epsilon$-neighborhood of training set $N_{\epsilon}=\bigcup_{s_i\in S_d}B(s_i,\epsilon)$ and data manifold $\mathcal{Z}$. $M$ is the upper bound of loss of data manifold $\mathcal{Z}$.

\label{lem_d_final}
\end{thm}

Similarly, the generalizability of the generator for RGAN can be defined as the convergence of  $|Q^\phi_m-Q^\phi|$, where $Q^\phi=\min_{G}V(G,D^*)$ and $Q^\phi_m=\min_{G}V_m(G,D^*)$. For original GANs, the generalizability of the generator is defined by $|U^\phi_m-U^\phi|$, where $U^\phi=\min_{G}S(G,D^*)$ and $U^\phi_m=\min_{G}S_m(G,D^*)$.  We can also prove that the RGAN's generator has a tighter upper bound of the generalizability of the generator than traditional GANs.

\begin{thm}
If the training set $S_g$ for generator consists of $n$ i.i.d samples, the generator of our proposed method $G_r$ and original GAN $G_{org}$ are $(K,\epsilon(S_g))$ robust, $G_r$ is $(\epsilon',S_g,\sigma'_{r})$ adversarial robust and $G_{org}$ is $(\epsilon',S_g,\sigma'_{org})$ adversarial robust, then, for any $\delta>0$ and small enough $\epsilon'$, with the probability at least $1-\delta$,  we have
\begin{eqnarray*}
\footnotesize{
\begin{aligned}
|Q^\phi_m-Q^\phi|\leq \gamma'_1 \sigma'_r + (1-\gamma'_1)\epsilon(S_g) + M'\sqrt{\frac{2Kln2+2ln\frac{1}{\delta}}{n}}\\
|U^\phi_m-U^\phi|\leq \gamma'_2 \sigma'_{org} + (1-\gamma'_2)\epsilon(S_g) + M'\sqrt{\frac{2Kln2+2ln\frac{1}{\delta}}{n}}
\end{aligned}
}
\end{eqnarray*}
$G_r$ obtains the tighter upper bound, i.e.:
\begin{eqnarray*}
&\gamma'_1 \sigma'_r + (1-\gamma'_1)\epsilon(S_g) + M'\sqrt{\frac{2Kln2+2ln\frac{1}{\delta}}{n}}\\
&\leq \gamma'_2 \sigma'_{org} + (1-\gamma'_2)\epsilon(S_g) + M'\sqrt{\frac{2Kln2+2ln\frac{1}{\delta}}{n}}
\end{eqnarray*}
where $\gamma'_1,\gamma'_2\in [0,1]$, which are closely related to $\sigma'_r, \sigma'_{org}$ and the intersection of $\epsilon'$-neighborhood of the training set $N_{\epsilon'}=\bigcup_{s_i\in S_g}B(s_i,\epsilon')$ and data manifold $\mathcal{Z}'$. $M'$ is the upper bound of loss of data manifold $\mathcal{Z}'$.

\label{lem_g_final}
\end{thm}

It is noted that  Theorem~\ref{lem_d_final} and~\ref{lem_g_final} made very mild assumptions to achieve tighter bound for $G$ and $D$ with the notion of $(\epsilon,S, \sigma)$ adversarial robustness, which helps bridge model generalizability and adversarial robustness. Specifically, it describes the variation of loss by upper bound $\sigma$ on the $\epsilon$-neighborhood of training set $N_{\epsilon}=\bigcup_{s_i\in S_d}B(s_i,\epsilon)$. If $\epsilon$-neighborhood $N_\epsilon$ intersects with data manifold $\mathcal{Z}$ and the variation of loss is bounded by $\sigma$ in $N_\epsilon$, then the loss of intersection region can be also bounded by $\sigma$. Therefore, the generalizability on the intersection region can be bounded. 
When we set a small enough radius $\epsilon$ (as we did in practice),  $B(s_i,\epsilon)$ does not contain data points with different label of $s_i$ (for $s_i\in S$). In this case, the upper bound of our method would be guaranteed tighter than original GANs. Some works have investigated the relationship between the adversarial robustness and generaliability~\cite{max_margin_network}. However, no exact bound has been obtained.


\section{Experiments}
We present a series of experiments in this section. First, we show that our proposed RGAN can improve the performance of different types of baseline models including WGAN-GP, DCGAN, WGAN-GP (resnet), and BWGAN. Inception score and FID are used to evaluate the quality of generations. 


\begin{table*}[h!]
\centering
\caption{Generation performance of different models on CIFAR-10 and STL-10 (without feeding the label information)}
\label{per_table}
\scriptsize
\begin{tabular}{ccccc}
\toprule
\multirow{2}{*}{Methods} & \multicolumn{2}{c}{Inception Score} & \multicolumn{2}{c}{FID}  \\
\cmidrule(r){2-3} \cmidrule(r){4-5}
&  CIFAR-10      &  STL-10
&  CIFAR-10      &  STL-10    \\
\midrule
Real data  & $11.24\pm0.12$ &$26.08\pm0.26$ & $7.8$ & $7.9$   \\
\midrule
Weight clipping            & $6.41\pm0.11$ &$7.57\pm0.10$ &$42.6$ &$64.2$                     \\
Layer norm              &$7.19\pm0.12$ &$7.61\pm0.12$ &$33.9$ &$75.6$           \\
Weight norm              &$6.84\pm0.07$ &$7.16\pm0.10$ &$34.7$ &$73.4$                      \\
Orthonormal            &$7.40\pm0.12$ &$8.56\pm0.07$ &$29.0$ &$46.7$ \\

ALI~\cite{warde2016improving}             &$5.34\pm 0.05$                          &                    &                    &                      \\
BEGAN~\cite{berthelot2017began}             &$5.62$                          &                     &                    &                     \\
DCGAN~\cite{radford2015unsupervised}             &$5.77\pm0.02$                          & $7.36\pm 0.06$                    & 42.18                   &  $53.23$                   \\
Improved GAN (-L+HA)~\cite{salimans2016improved}             &$6.86\pm0.06$                          &                     &                    &                       \\
EGAN-Ent-VI~\cite{dai2017calibrating}             &$7.07\pm 0.10$                          &                     &                    &                       \\
DFM~\cite{warde2016improving}             &$7.72\pm 0.13$                          &                     &                    &                       \\
CT GAN~\cite{wei2018improving}             &$8.12\pm 0.12$                          &                     &                    &                       \\
SNGAN~\cite{miyato2018spectral}             &$8.22\pm 0.05$                          & $9.10\pm 0.04$                    & 21.70                   &  $40.10$                     \\
BWGAN~\cite{adler2018banach}             &$8.08\pm 0.05$                          &                     & 25.67                   &                       \\
WGAN-GP-res~\cite{gulrajani2017improved}             &$7.76$                          & $9.06\pm0.03$                    & $22.19$                   &   $42.60$                    \\
\midrule
RGAN (WGAN-GP-res)             &$\textbf{8.25}\pm 0.1$                          & $\textbf{9.16}\pm 0.02$                    & $\textbf{19.79}$                   & $\textbf{39.62}$                      \\
\bottomrule
\end{tabular}
\end{table*}

\begin{table}[h!]
\centering
\caption{Performance of conditional models on CIFAR-10 (with feeding the label information)}
\label{per_table_super}
\scriptsize
\begin{tabular}{ccccc}
\toprule
\multirow{1}{*}{Methods} & \multicolumn{1}{c}{Inception Score} & \multicolumn{1}{c}{FID}  \\

\midrule
Real data  & $11.24\pm0.12$ &$26.08\pm0.26$    \\
\midrule
DCGAN            & $6.58$ &                     \\
AC-GAN              &$8.25\pm0.07$ &            \\
SGAN-no-joint              &$8.37\pm0.08$ &            \\
WGAN-GP-res              &$8.29\pm0.10$ &  $19.5$          \\
SNGAN              &$8.37\pm0.14$ &   $19.2$         \\
\midrule
RGAN (SNGAN)             &$\textbf{8.65}\pm 0.13$                          & $\textbf{17.2}$                                          \\
\bottomrule
\end{tabular}
\end{table}

\begin{figure*}[ht]
\centering
\subfigure[Inception score on CIFAR-10] {
\includegraphics[width=0.23\textwidth]{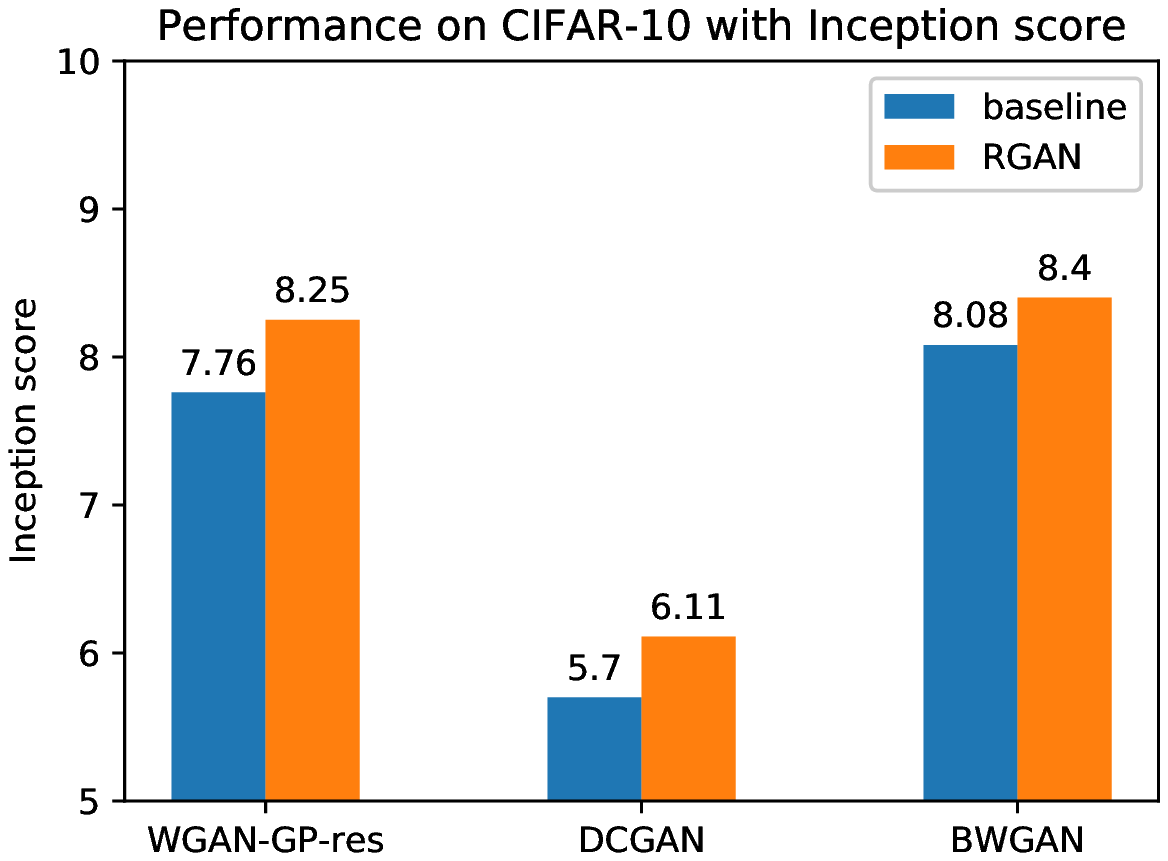}
\label{fig:ad1} }
\subfigure[FID on CIFAR-10] {
\includegraphics[width=0.23\textwidth]{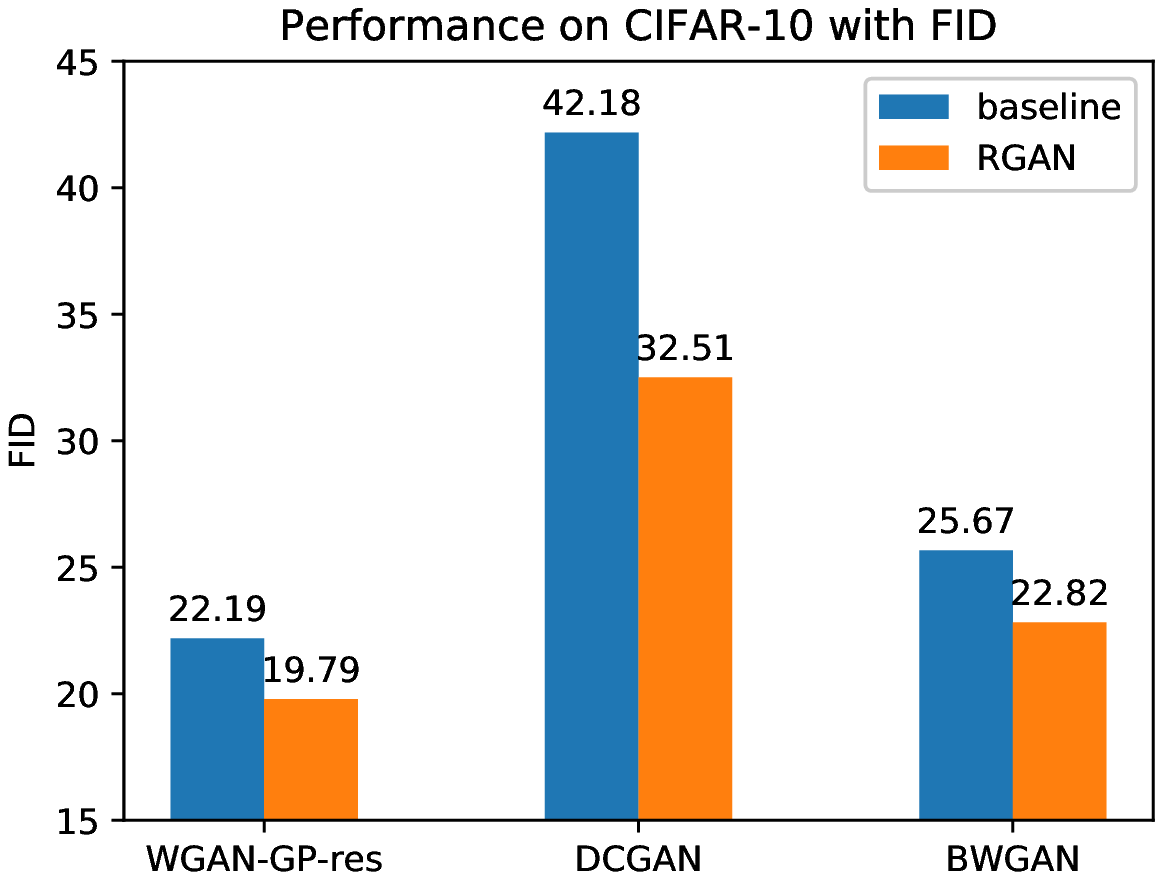}
\label{fig:ad2} }
\subfigure[Inception score on STL-10] {
\includegraphics[width=0.24\textwidth]{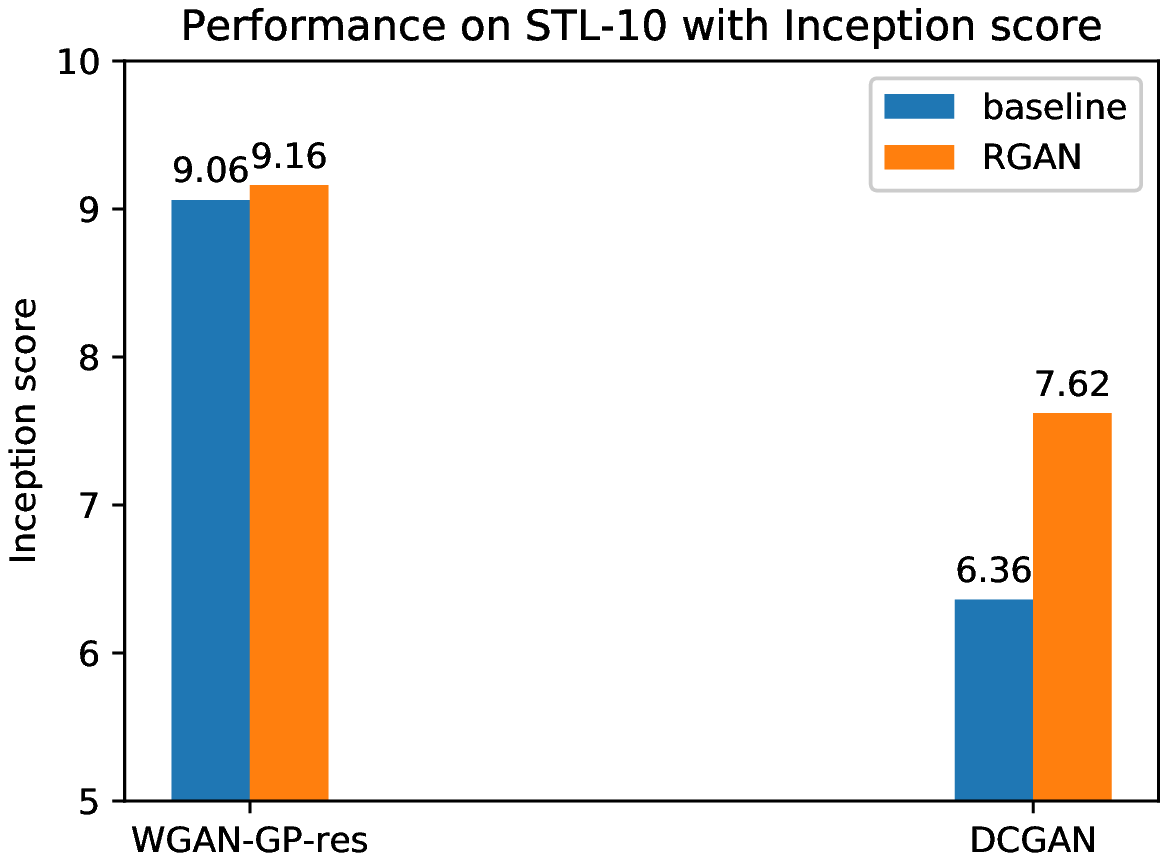}
\label{fig:ad3} }
\subfigure[FID on STL-10] {
\includegraphics[width=0.24\textwidth]{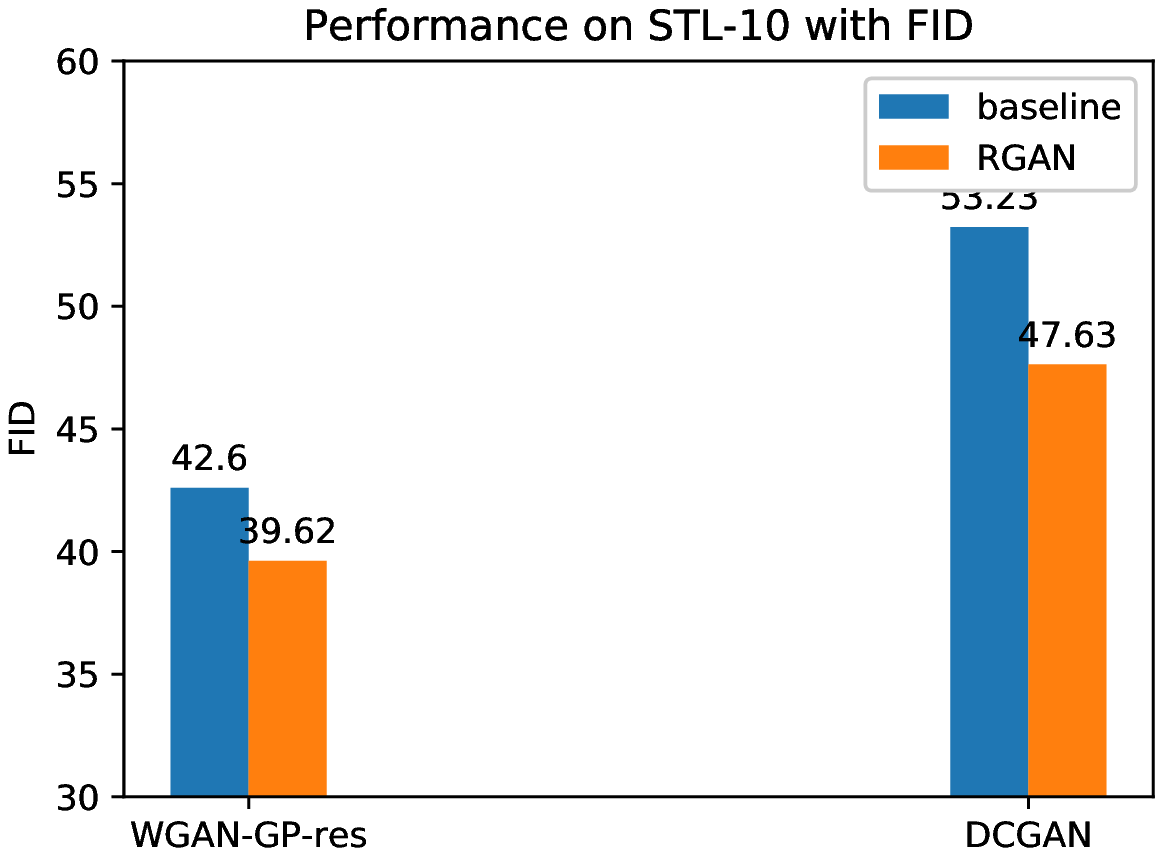}
\label{fig:ad4} }

\caption{Performance (Inception score: the bigger the better, and FID: the lower the better) of different baselines (blue bars) and corresponding RGANs (orange bar). Our methods consistently perform better than baselines on different datasets and criteria.$^2$}
\label{hist}
\label{tsnetext}
\end{figure*}

\begin{figure*}[h]
\centering
\subfigure[WGAN-GP vs RGAN] {
\includegraphics[width=0.23\textwidth]{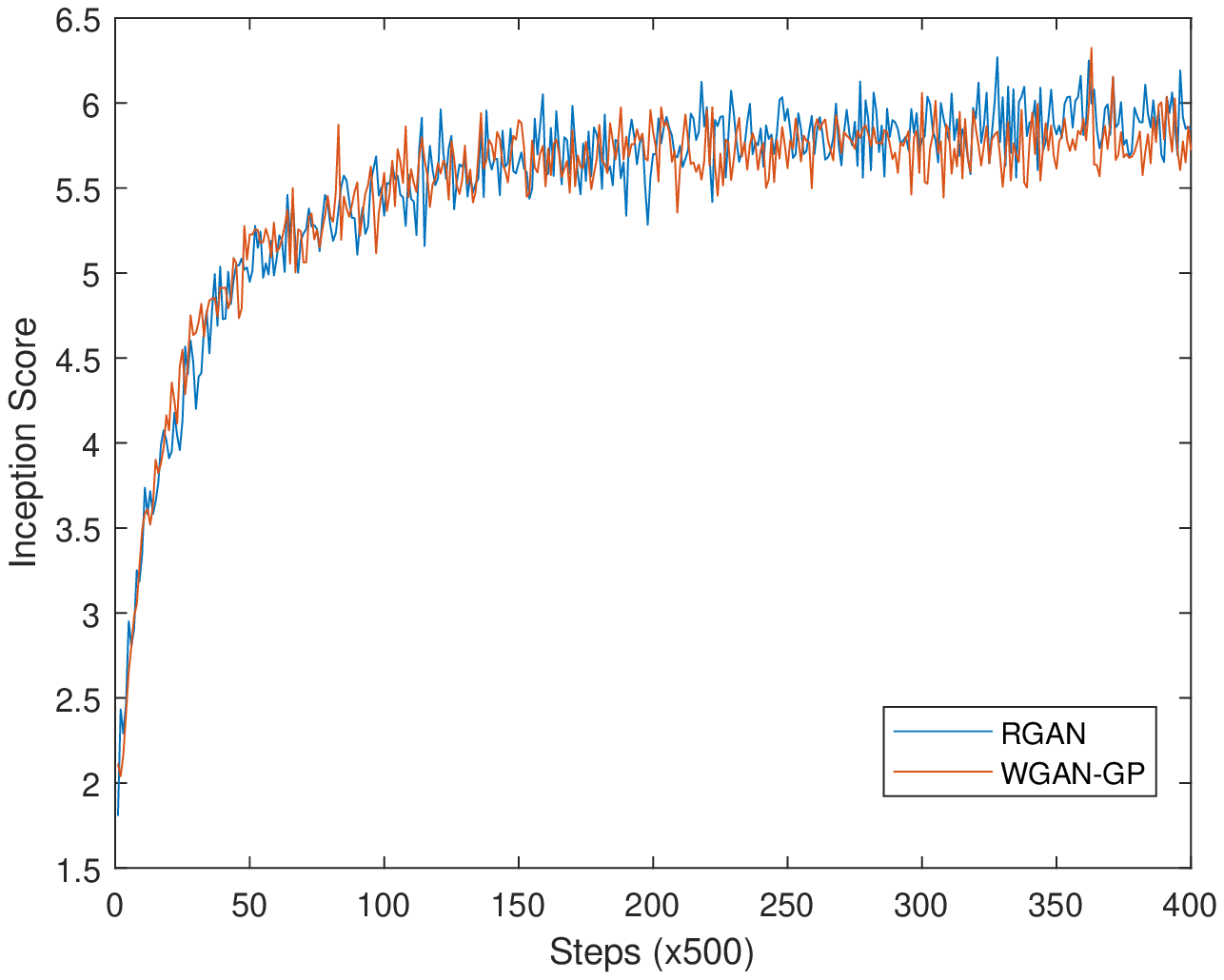}
\label{fig:ad1} }
\subfigure[WGAN-GP (res) vs RGAN] {
\includegraphics[width=0.23\textwidth]{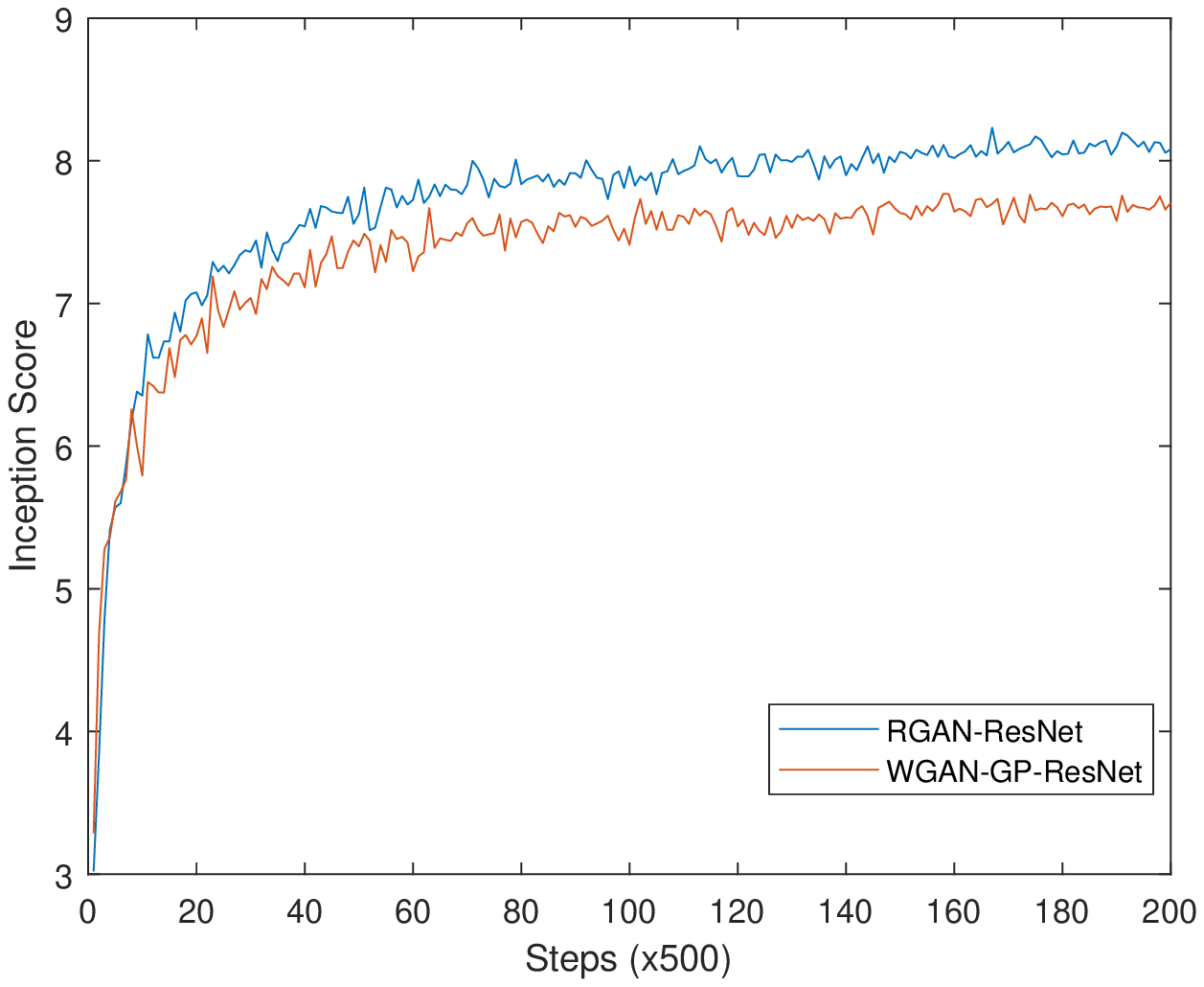}
\label{fig:ad2} }
\subfigure[DCGAN vs RGAN] {
\includegraphics[width=0.23\textwidth]{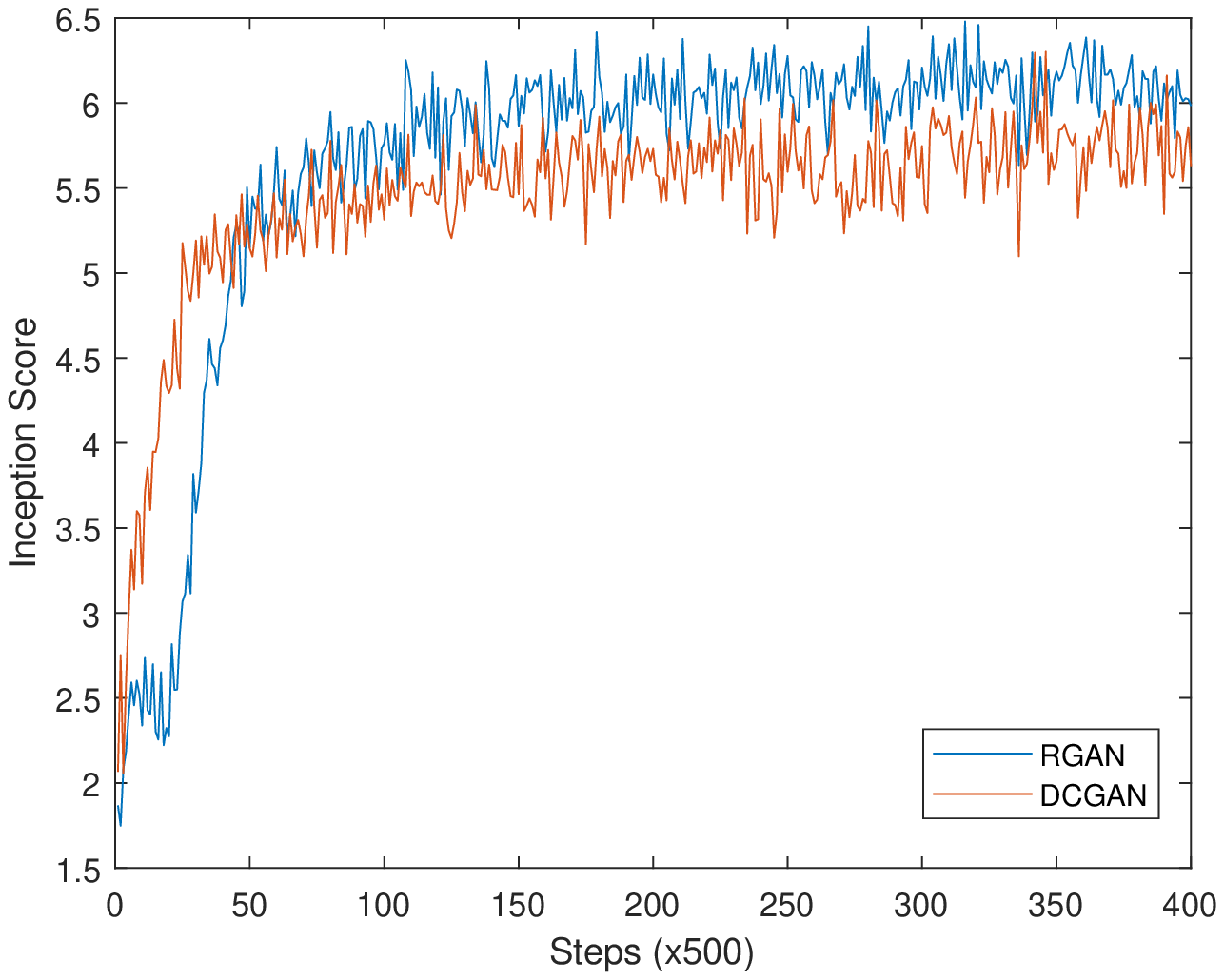}
\label{fig:ad3} }
\subfigure[BWGAN vs RGAN] {
\includegraphics[width=0.23\textwidth]{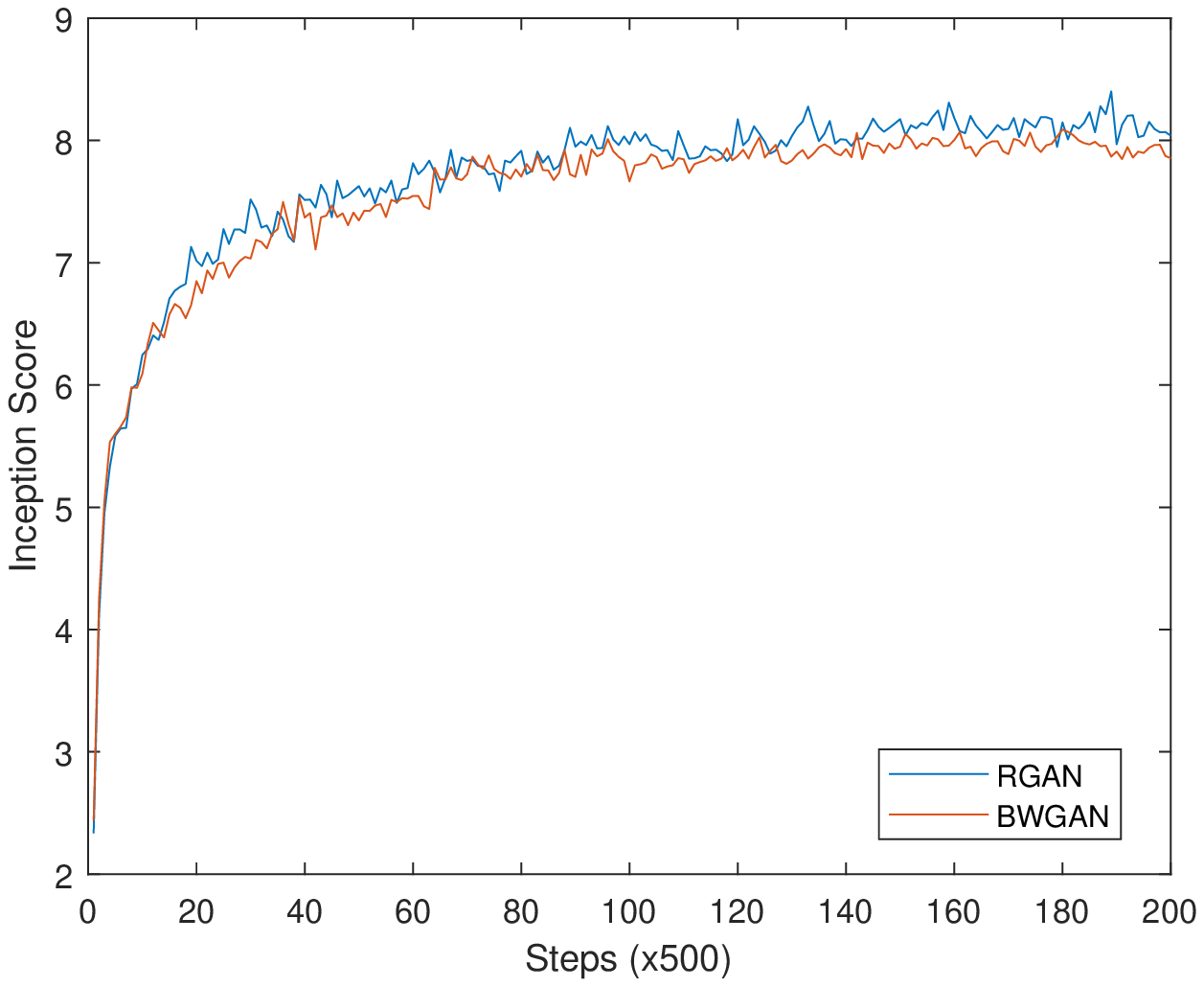}
\label{fig:ad4} }

\caption{Inception score  versus training step. Each subfigure shows the comparison between  a different baseline model (blue curve) and its corresponding robust version (by applying the RGAN strategies, red curve). Robust GANs consistently achieve much better performance though they  converge in a similar speed to baseline models.}
\label{fig_curve}
\label{tsnetext}
\end{figure*}

\begin{table*}[h!]
\caption{Ablation analysis for RGAN on different baselines on CIFAR-10.}
\label{fig_ablation}
\centering
\scriptsize
\begin{tabular}{l|ccccc}
\hline
           & WGAN-GP  & WGAN-GP (res) & DCGAN & BWGAN   \\ \hline
 Baseline (without robust training)                                       & $5.77\pm 0.02$ &  $7.76\pm 0.13$    &$5.70\pm 0.05$ &$8.08$                \\ \hline
 Baseline (with random noise)                                       & $5.82\pm 0.02$ &  $7.75\pm 0.11$    &$5.76\pm 0.06$ &$8.03$                \\ \hline
 Robust training on generator only              & $5.89\pm 0.02$ &  $7.86\pm 0.13$   &$5.80\pm 0.02$ &$8.11$           \\ \hline
 Robust training on discriminator only              & $5.87\pm 0.03$ &  $8.01\pm 0.11$   &$6.02\pm 0.02$ &$8.23$           \\ \hline
 Robust training on generator \& discriminator & $\textbf{5.91}\pm 0.02$ &  $\textbf{8.25}\pm 0.10$    &$\textbf{6.11}\pm 0.02$ &$\textbf{8.40}$           \\ \hline
\end{tabular}
\end{table*}







\subsection{Quantitative Comparison}
To evaluate the performance  of RGAN, we follow the previous works~\cite{gulrajani2017improved,adler2018banach} on robustness and mainly conduct experiments on  CIFAR-10 and STL-10. In addition to other existing popular GAN models, we focus on comparing $5$ models including WGAN-GP, WGAN-GP (resnet), DCGAN, SNGAN, and BWGAN, all of which try to control the discriminator. We implemented our proposed robust strategy on these baselines and would like to check if the robust training could indeed improve the performance. The structures and settings of our method are the same as baseline models. We train WGAN-GP, DCGAN, BWGAN and our proposed RGANs with $50,000$ training samples for $200,000$ epochs. For WGAN-GP (resnet) and our corresponding model, $100,000$ epochs appear sufficient. For each $500$ epochs, we calculate the inception score for $50,000$ generated images. For training RGANs, there are three hyper-parameters $\lambda$ (used to trade-off our objective and the original one), $\epsilon_1$ and $\epsilon_2$. We set $\lambda=0.1$ which was searched from $\{0.001, 0.01, 0.1, 0.5, 1, 2\}$. We also set $\epsilon_1=0.01$ and $\epsilon_2=4$ which was searched from $\{0.001, 0.01, 0.1, 0.2, 0.5, 1, 2, 4, 5, 10\}$. For STL-10, we train our models and corresponding baselines with $800K$ training samples. The training settings keep the same as those for CIFAR-10. 

 We list the performance for different models in Table~\ref{per_table}. Clearly, our proposed RGAN (which is based on WGAN-GP-res) achieves the best result among all the methods in terms of both the criteria, i.e., Inception Score and FID. In order to check if the proposed robust strategy can indeed improve over different baselines, we also detail the performance in Fig.~\ref{hist} where we plot the bar charts for different baseline models and their robust version with RGAN on two datasets (CIFAR-10 and STL-10).\footnote{BWGAN appears not to converge in STL-10 in our experiments. For fair comparison, we did not report the performance when BWGAN is used as the baseline in STL-10.} It is noted that the robust strategy can consistently improve the baselines on the two datasets in terms of both the criteria. In addition, we also show the convergence curves in Figure~\ref{fig_curve}. Clearly, when our robust strategy is applied on the baseline GANs, an obvious increase of the inception scores can be observed (though the convergence speed is similar to that of baseline models).

 We also check if our robust strategy can boost the performance of the conditional versions of various GANs by taking CIFAR-10 as one illustrative example when the class or label information can be feeding in training. These results are reported in Table~\ref{per_table_super} where SNGAN was adopted as the baseline to implement RGAN. Evidently, the RGAN again demonstrates the highest performance, significantly better than the other models.

\begin{figure*}[h]
\begin{minipage}[t]{0.73\linewidth}
\centering
\subfigure[WG.-GP-res] {
\includegraphics[width=0.25\textwidth]{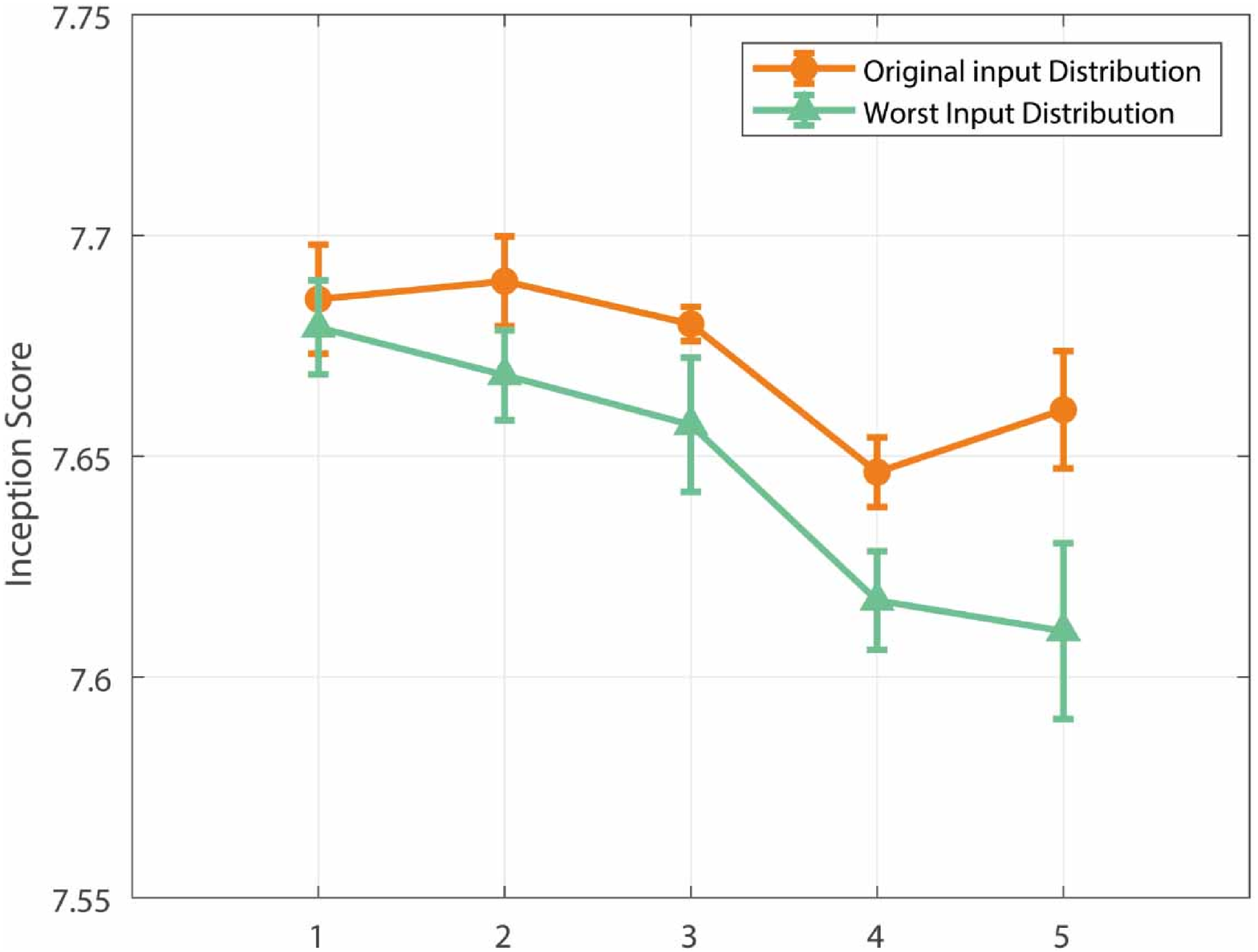}
\label{fig:ad1} }
\subfigure[DCGAN] {
\includegraphics[width=0.25\textwidth]{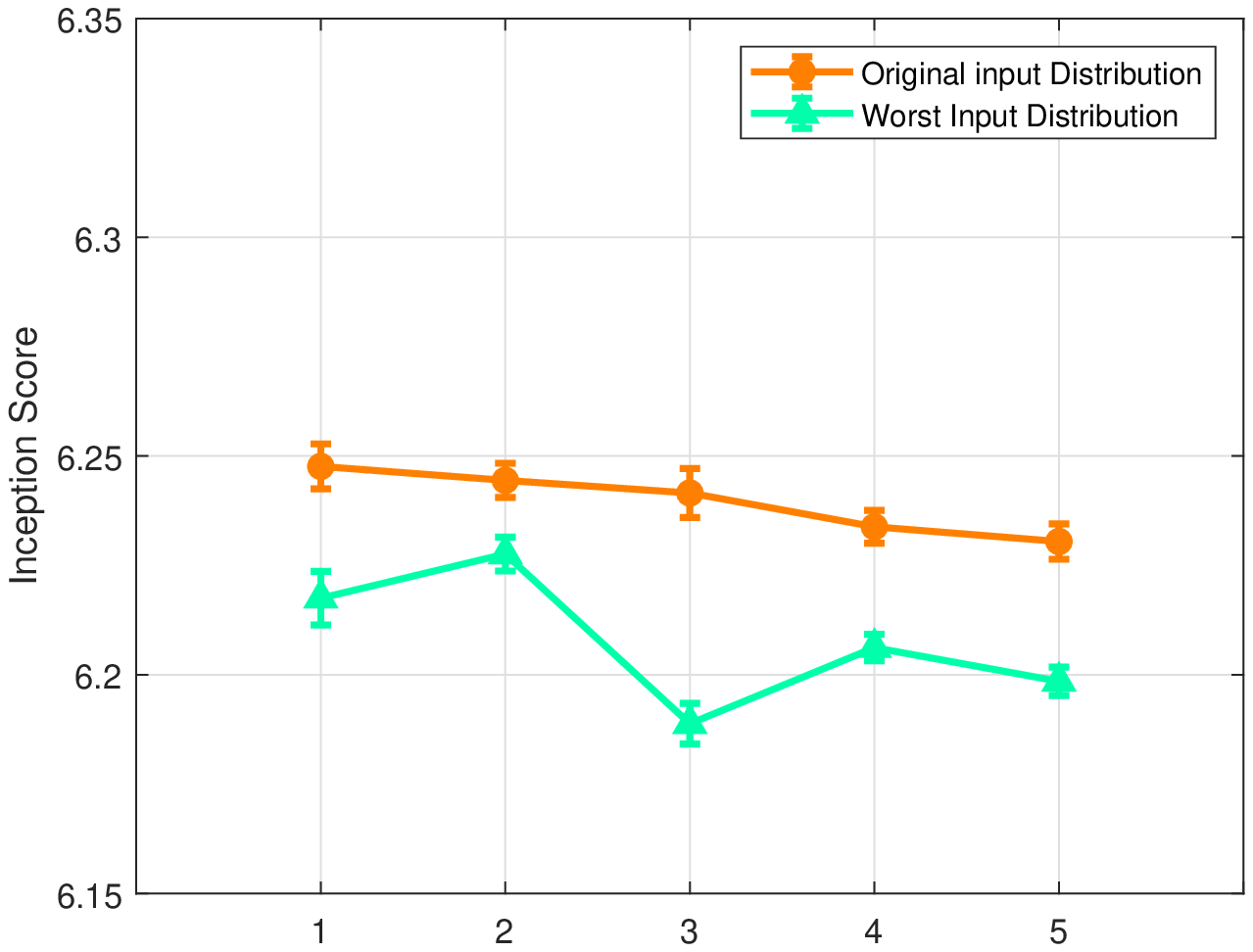}
\label{fig:ad1} }
\subfigure[WGAN-GP] {
\includegraphics[width=0.25\textwidth]{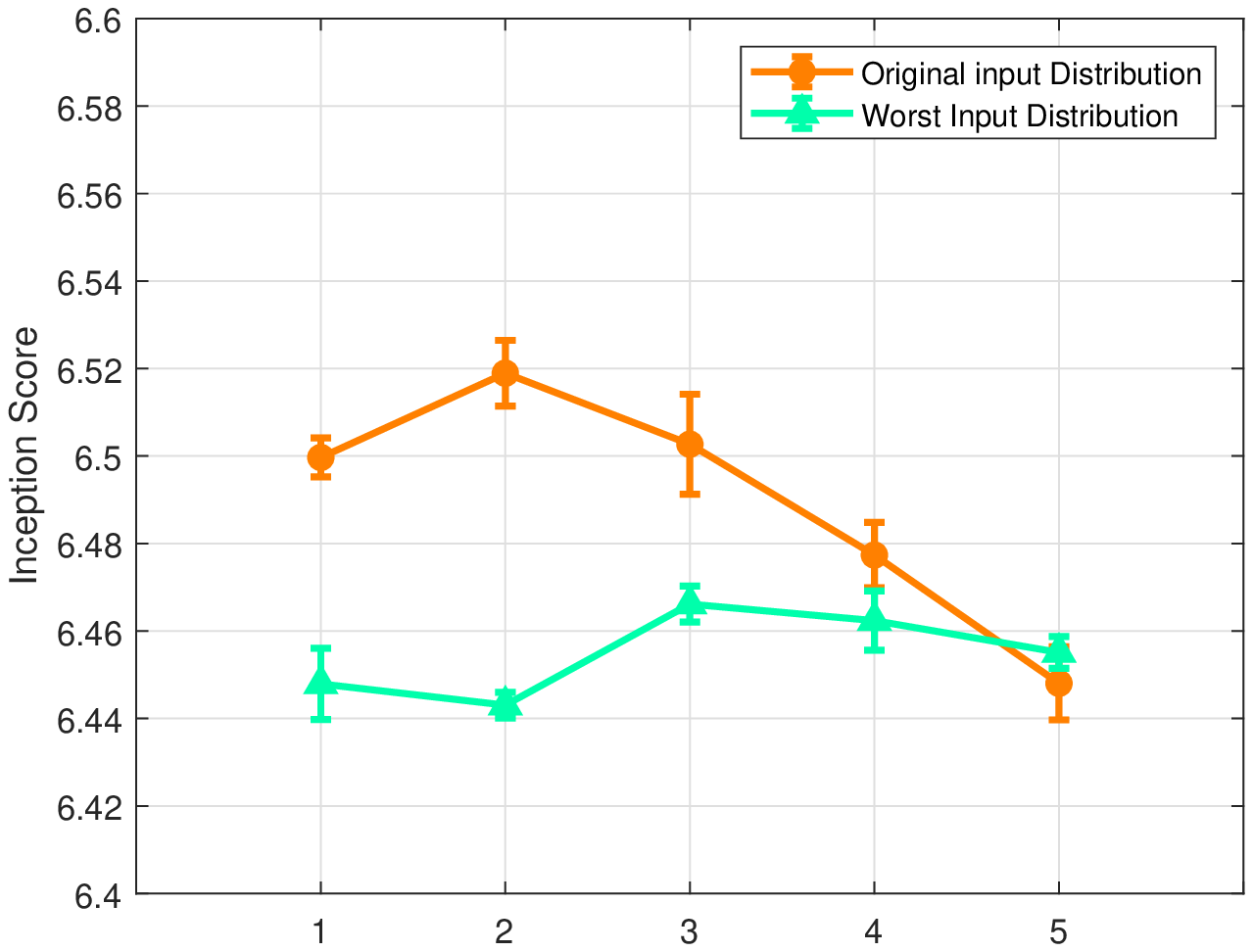}
\label{fig:ad1} }\\
\subfigure[ RGAN(WG.-GP-res)] {
\includegraphics[width=0.25\textwidth]{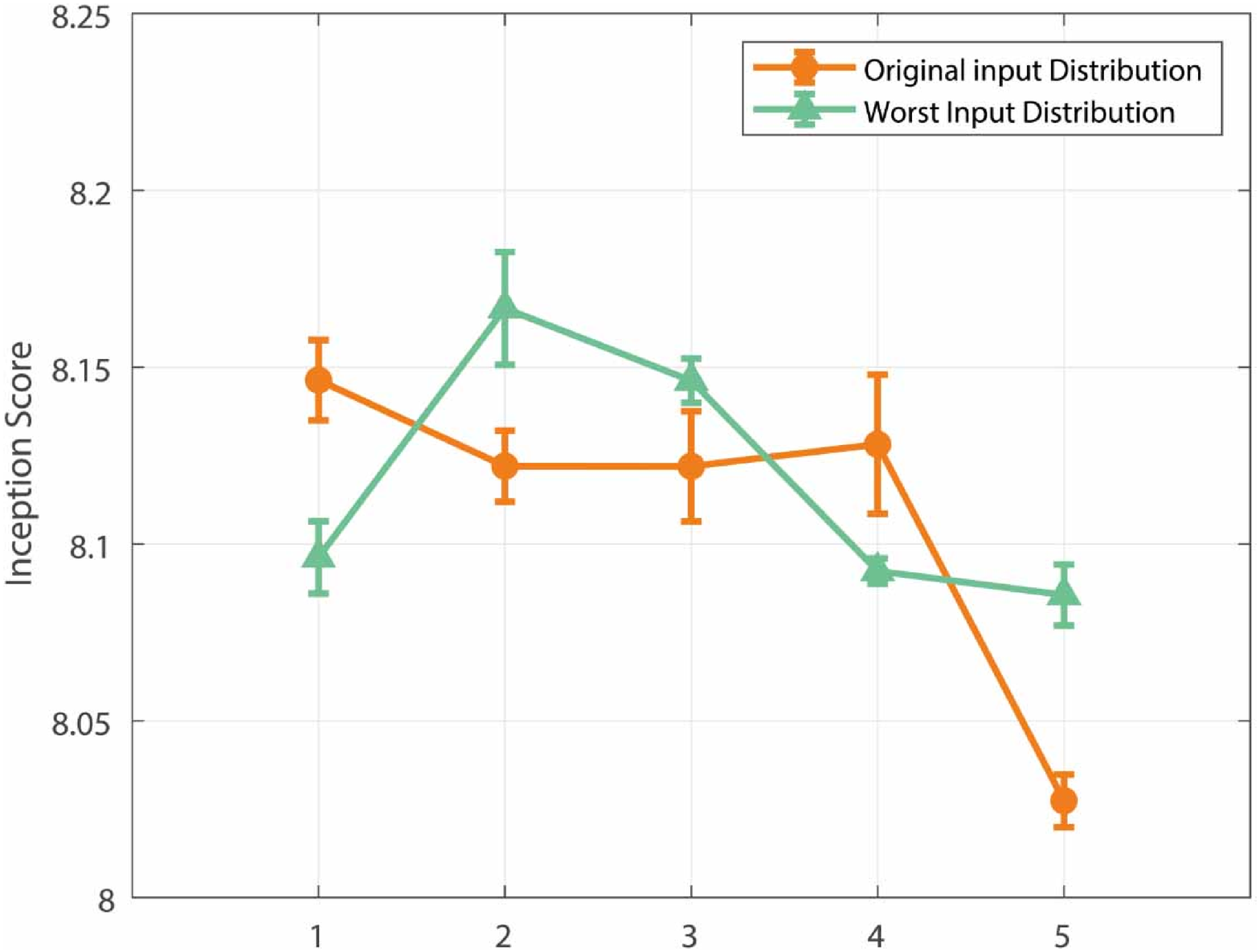}
\label{fig_per1} }
\subfigure[RGAN(DCGAN)] {
\includegraphics[width=0.25\textwidth]{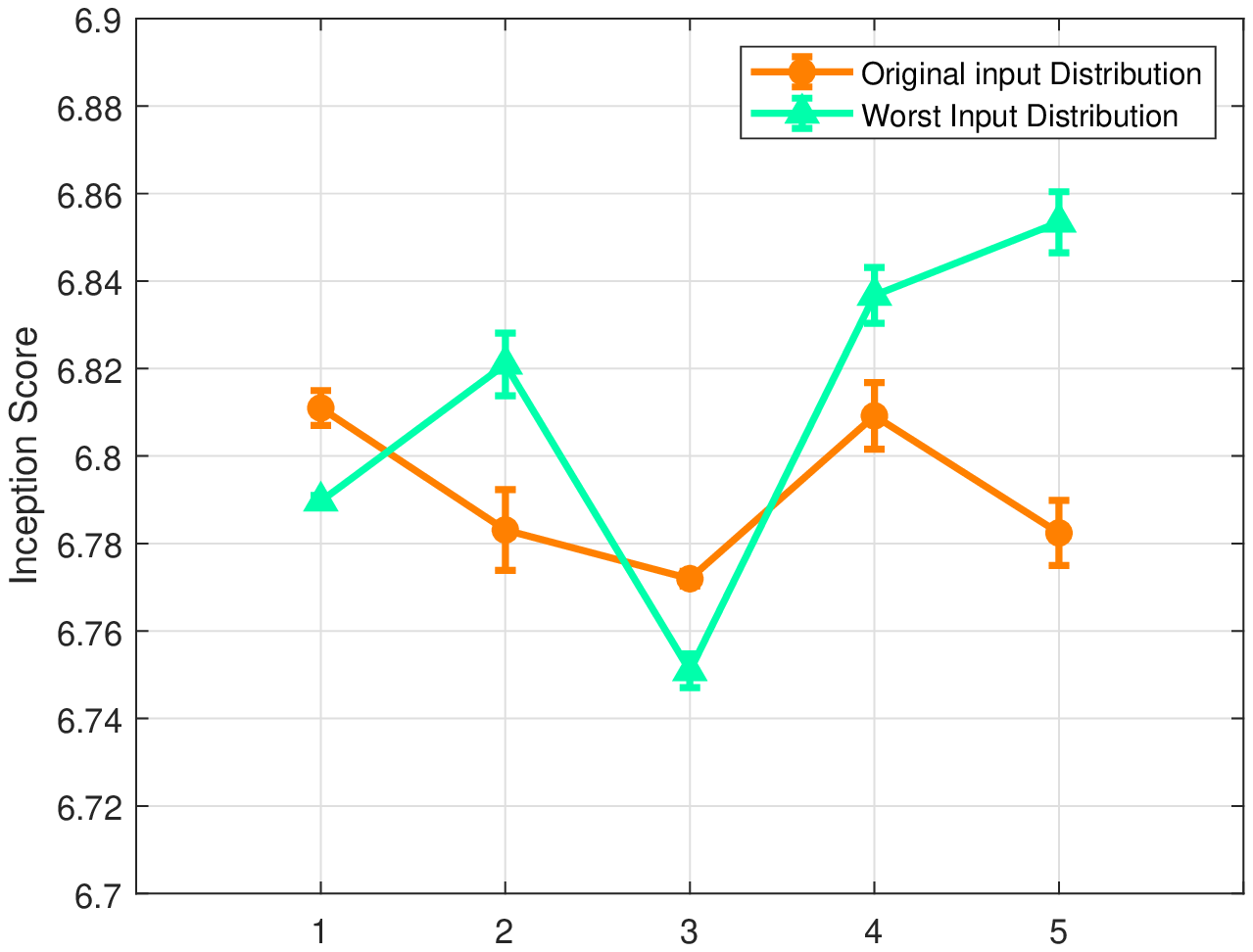}
\label{fig_per1} }
\subfigure[RGAN(WGAN-GP)] {
\includegraphics[width=0.25\textwidth]{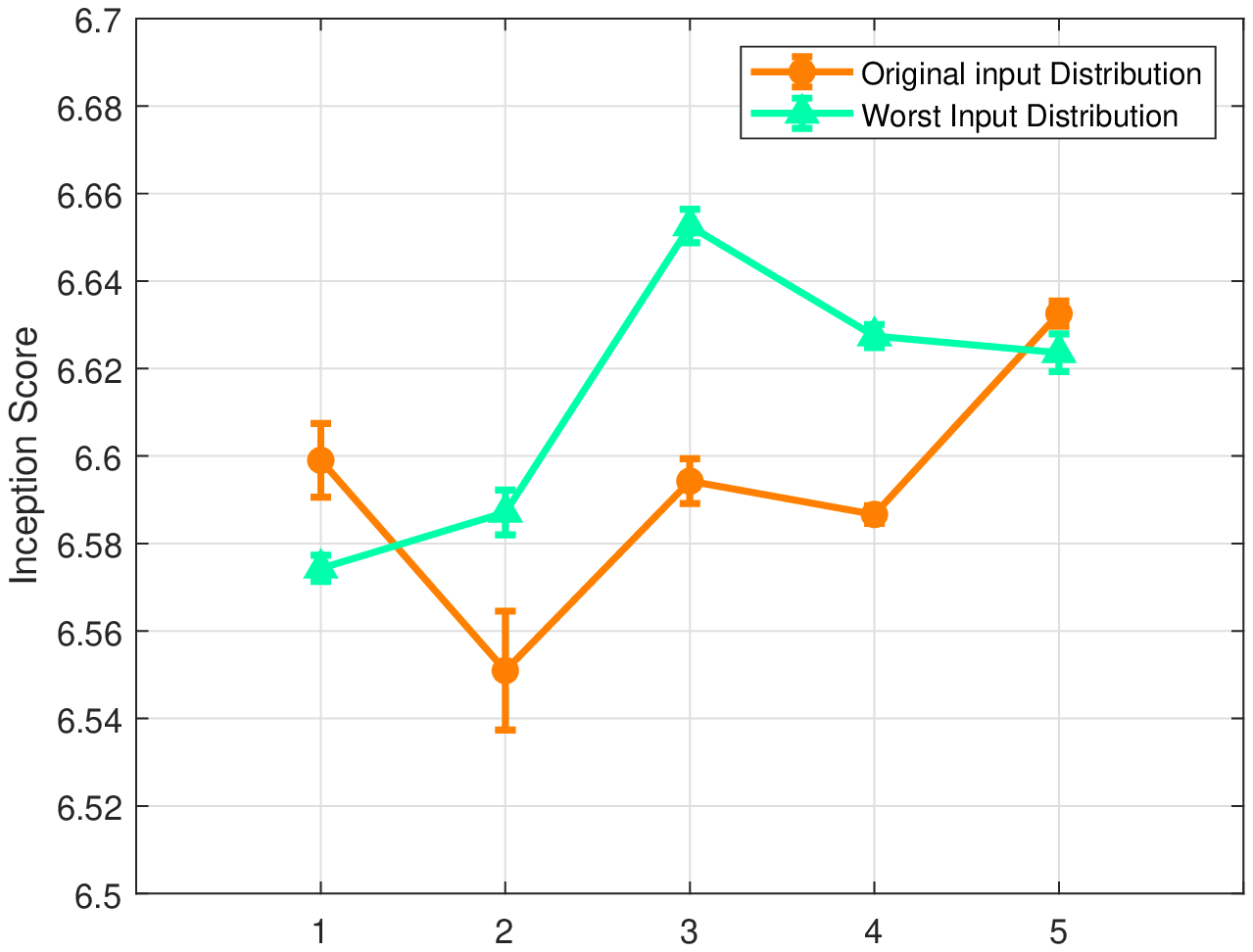}
\label{fig_per1} }
\caption{Inception score of baselines and RGANs on both the original input noise and the worst input noise on CIFAR-10. Performance of baselines are almost consistently degraded in the worst input noise (compared from the original input noise), while their robust versions (trained with RGAN) perform similar and stable for both worst and original input noise.}
\label{fig_per}
\label{tsnetext}
\end{minipage}
\begin{minipage}[t]{0.23\linewidth}
\subfigure[Perfor. vs.  $\epsilon_1$ in $G$] {
\includegraphics[width=0.75\textwidth]{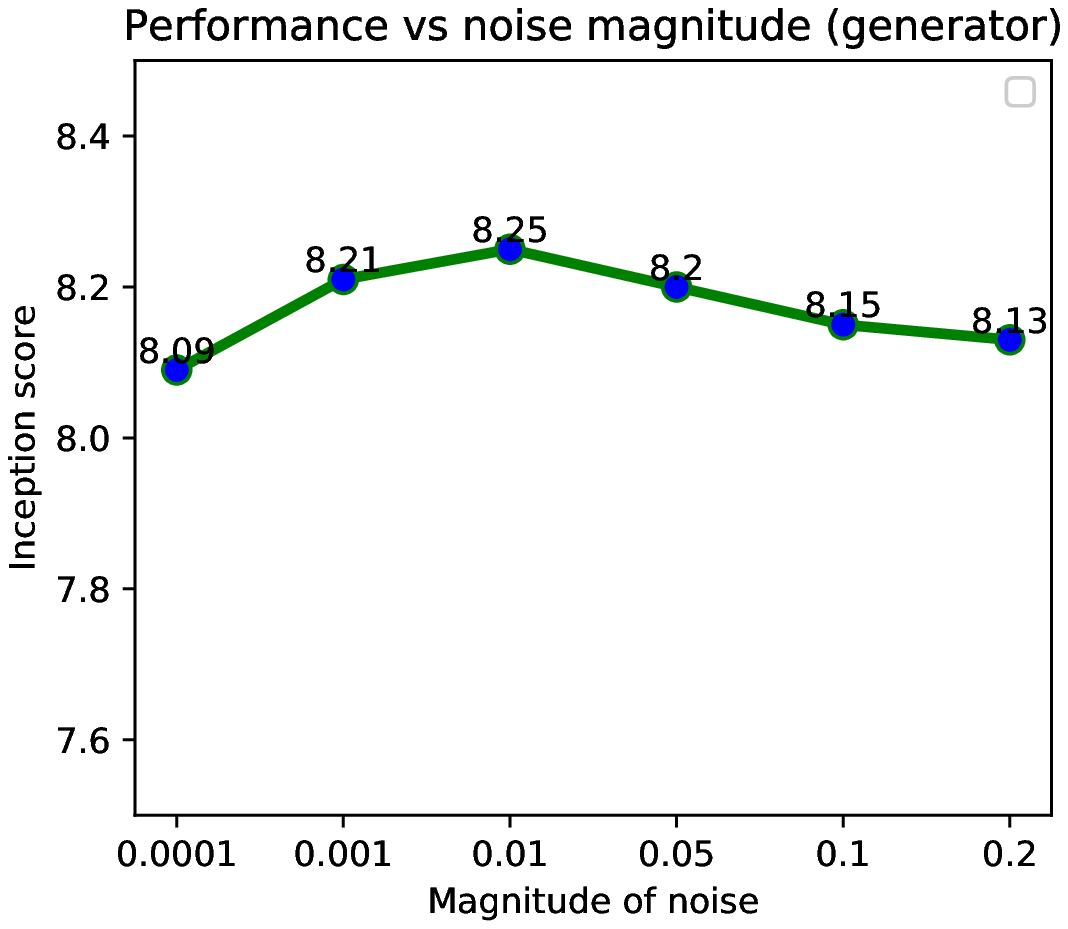}
\label{fig:sen_b} }\\
\subfigure[Perfor. vs.  $\epsilon_2$ in $D$] {
\includegraphics[width=0.75\textwidth]{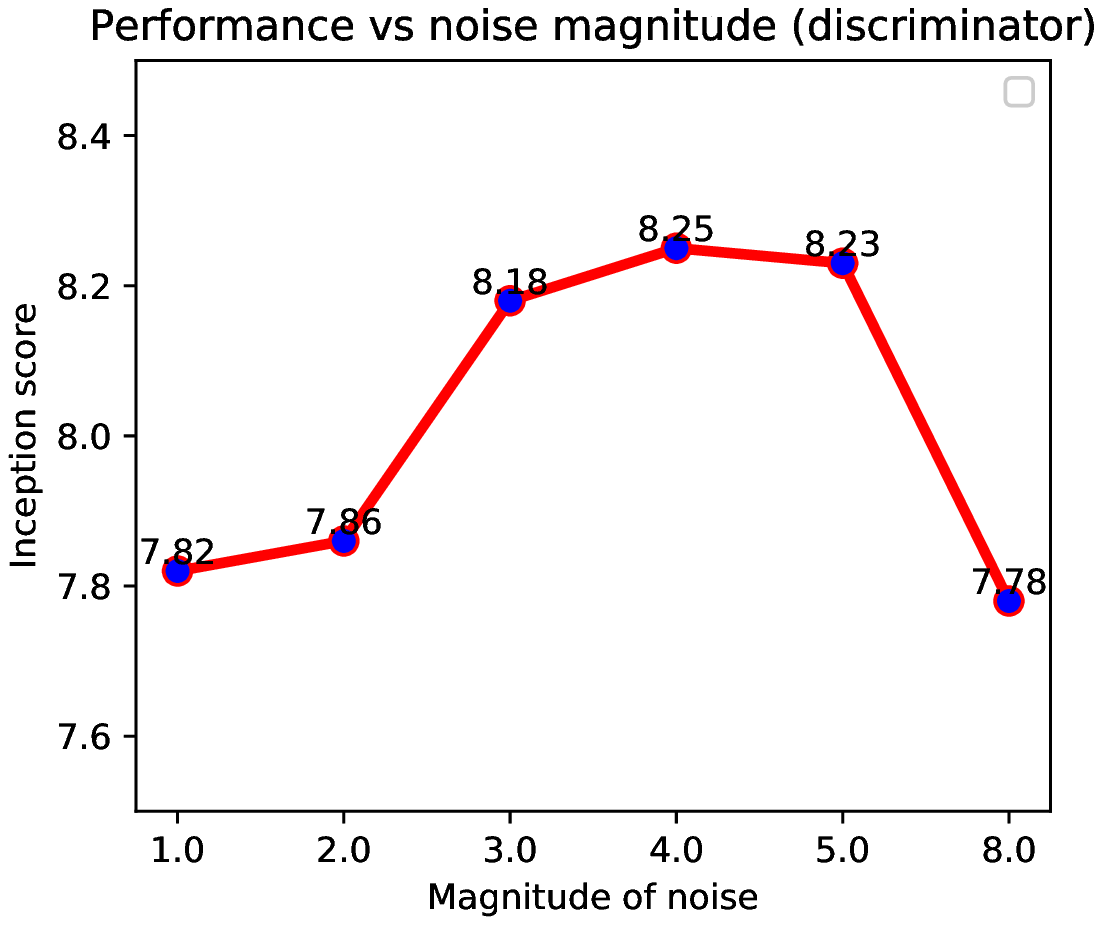}
\label{fig:per_n_d} }

\caption{Inception score (RGAN) vs. perturbation magnitude of noise for $G$ and $D$  on CIFAR-10. }
\label{fig:sen}
\end{minipage}
\end{figure*}

\begin{figure*}[h]
\centering
\subfigure[Input noise] {
\includegraphics[width=0.19\textwidth]{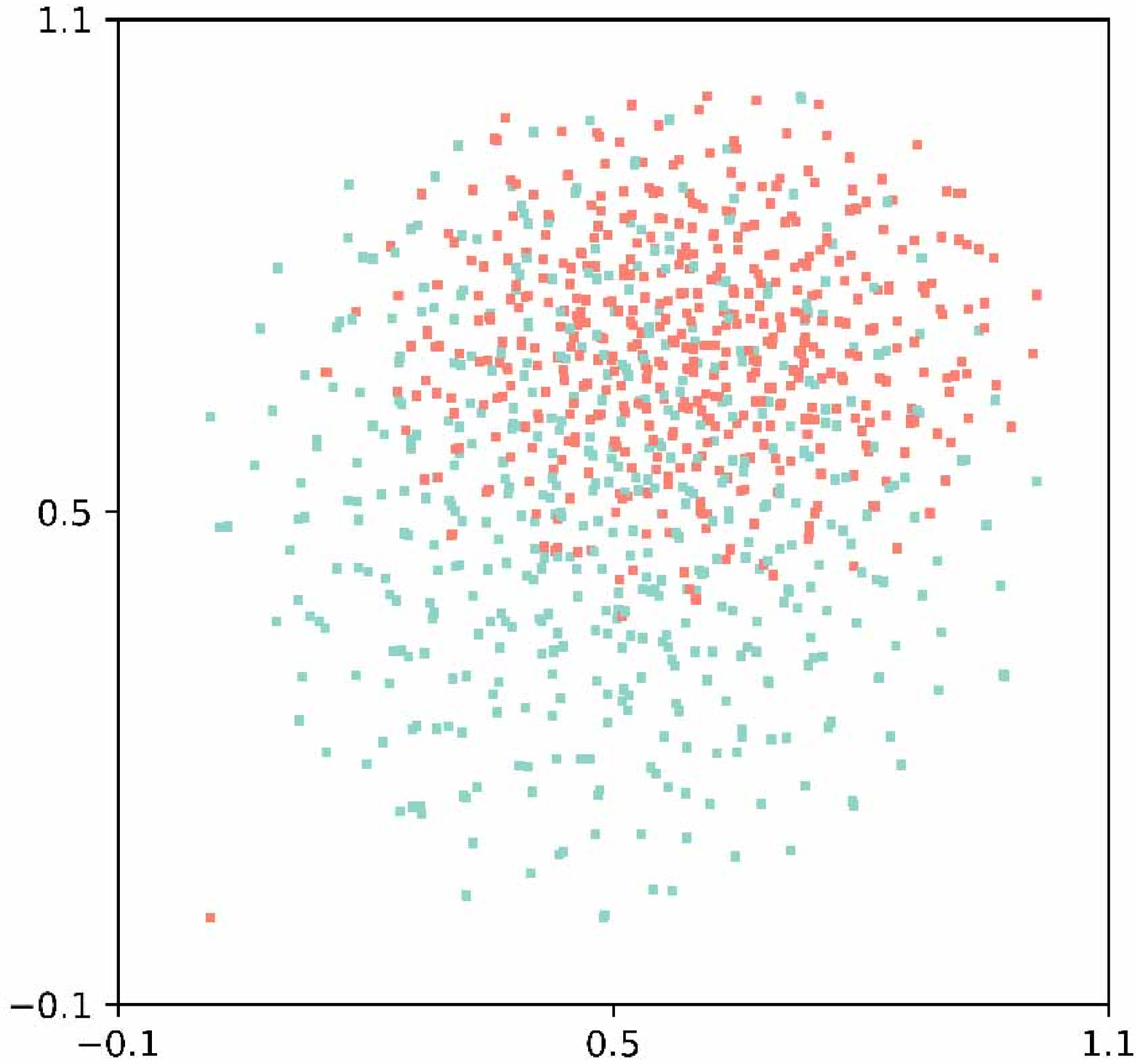}
\label{fig:ad1} }
\subfigure[Worst distribution] {
\includegraphics[width=0.19\textwidth]{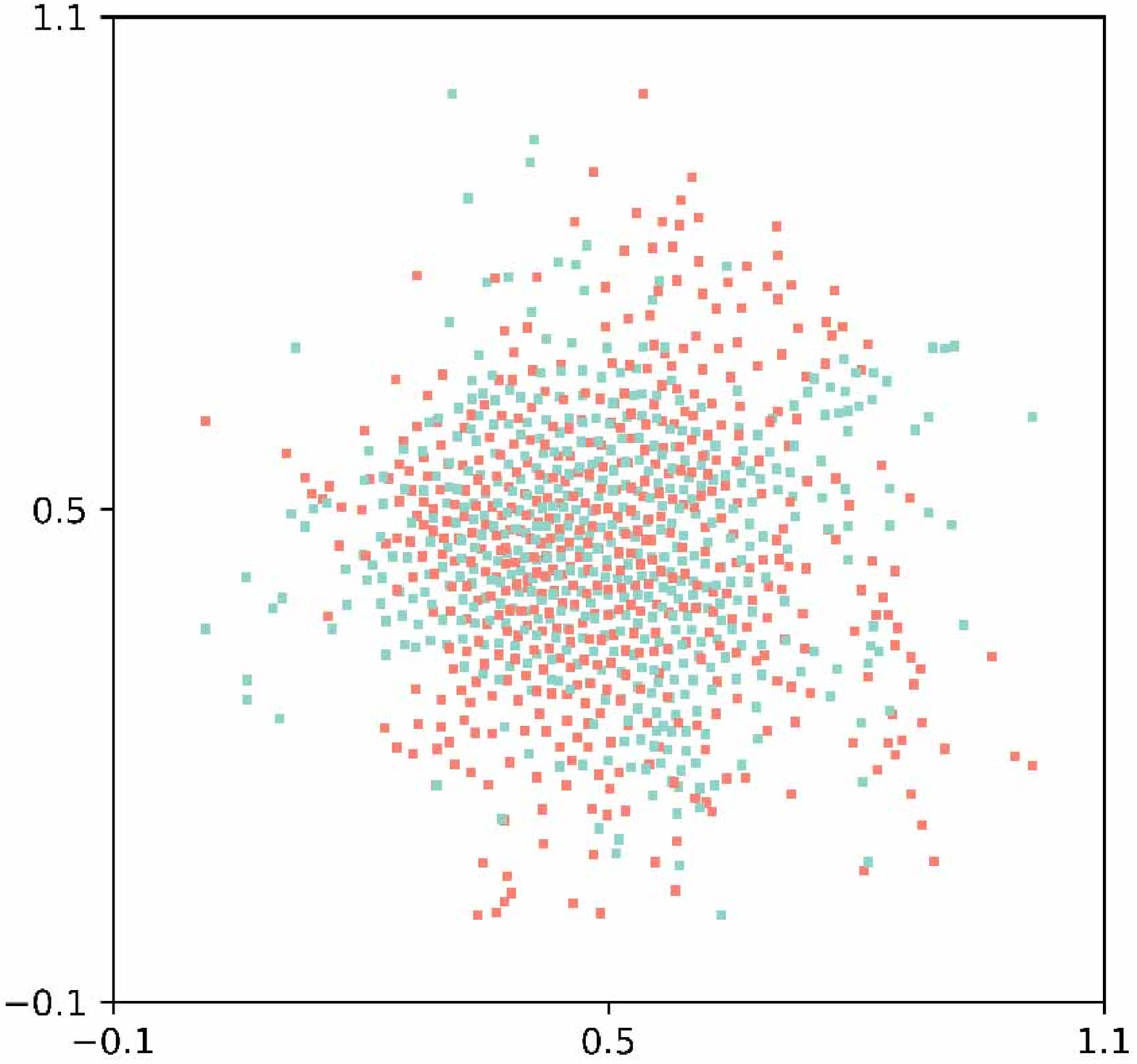}
\label{fig:ad3} }
\subfigure[Data by WGAN-GP] {
\includegraphics[width=0.19\textwidth]{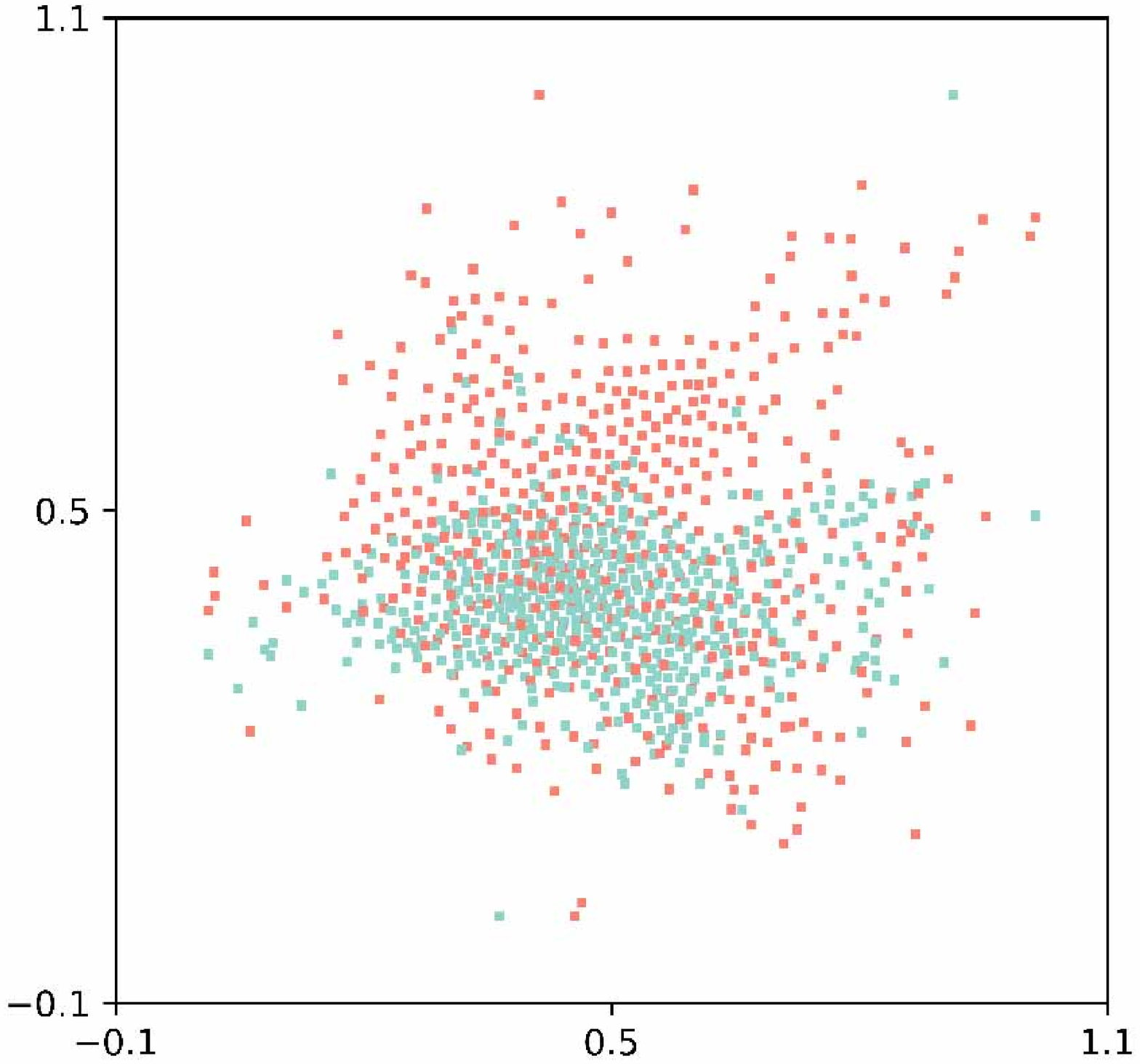}
\label{fig:ad4} }
\subfigure[Data by RGAN] {
\includegraphics[width=0.19\textwidth]{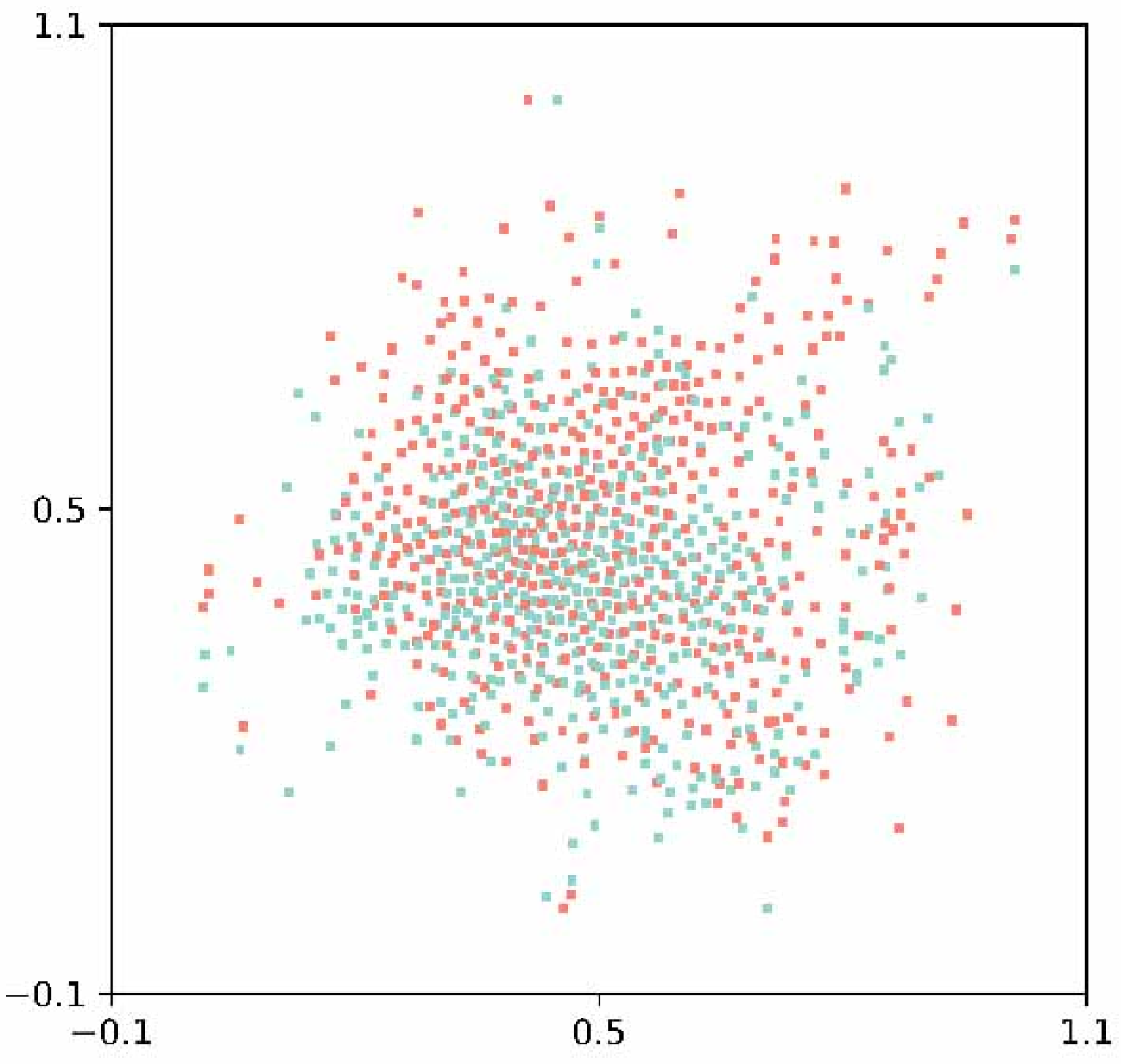}
\label{fig:ad5} }
\caption{Visualization and T-SNE embedding on CIFAR-10. \textbf{(a)}: Red  and blue points are  the input noise sampled respectively from the original Gaussian distribution and the worst distribution. The worst distribution covers a wider range of area, especially low density area of original distribution which might cause poor generation.  \textbf{(b)}: The worst real distribution (red) and worst generation distribution (blue). It is noted that the worst data distributions are more similar to each other which are more difficult to be classified.
\textbf{(c)}: Red and blue points are images sampled  from real distribution, and generated by WGAN-GP respectively. \textbf{(d)}: Red points are  sampled from real distribution, and blue points  are generated by RGAN. The data distribution generated by RGAN is apparently closer to the real distribution.
}
\label{tsnevis}
\end{figure*}
\subsection{Ablation and Sensitivity Analysis}
We conduct the ablation and sensitivity analysis in this subsection. Specifically, we experiment on CIFAR-10 with  robust training over generator only,  robust training over discriminator only, and robust training over both the generator and discriminator,  trying to see if  a robust training is necessary on both generator and discriminator. The results are listed   in Table~\ref{fig_ablation}. As observed, robust training on either generator or discriminator can consistently improve the performance of all baseline models, while a joint robust training on both generator and  discriminator can further boost the performance. We also experiment the baseline models which were injected with random Gaussian input noise on both discriminator and generator (at the same level of noise magnitude as used in RGAN). Though marginal improvement can be observed after noise injections, our robust training on either generator or discriminator or both shows consistently better performance. It is interesting to note that robust training on discriminator only could lead to more performance gain than on generator only, implying that a robust discriminator may be more important. This would be investigated as future work.

In addition, taking again CIFAR-10 as one illustrative dataset, we also show that our proposed robust method can perform robust on some potential input noise which might lead to poor generations (input noise sampled from the worst input distribution).  Specifically, we generate $50,000$ images with RGAN and various baseline models from the original distribution and worst distribution for five times. Then, we compute the inception score and their corresponding standard deviation. The results are showed in Fig.~\ref{fig_per}. As observed, without the robust training, those baseline models  perform consistently  worst in the case of the worst noises input  than that of the original input noises. This shows that the traditional GANs may not be robust and may lead to worse performance in case of certain poor input noise. In comparison, when the robust training is implemented, RGAN leads to similar performance even if the worst input noise is given.

Finally, we examine in Fig.~\ref{fig:sen} how the performance of RGAN would vary versus the two parameters $\epsilon_1$ and $\epsilon_2$ (see Algorithm 1) (used to control the perturbation magnitude in robust training) on CIFAR-10. Clearly, RGAN appears more sensitive to  the perturbation magnitude $\epsilon_2$ in discriminator, while it is less sensitive to the perturbation $\epsilon_1$ in generator.

\subsection{Visualization}
We present visualization results to compare various methods qualitatively on CIFAR-10 and CelebA.
We  first present a series of visualization trying to understand visually why the robust GAN could lead to better performance than the traditional GANs. To this end, we sample $500$ data points from the original input distribution and worst input distribution respectively. We then plot the 2-dimensional T-SNE embedding of these points. We also would like to plot the real data distribution and the generated data from the traditional GAN as well as our robust GAN. For clarity, we take WGAN-GP as one example but we should bear in mind that the conclusion is basically the same for other traditional GANs like DCGAN. These plots are made in Figure~\ref{tsnevis} where one can inspect the meaning of each subfigure in the caption. We highlight some remarks as follows. First, Figure~\ref{tsnevis}(a) indicates that the worst distribution covers wider range of areas, especially low density areas of the original distribution; this might cause poor generations since the worst input noise distribution is significantly different from the original input noise.  Second, (b) shows that the worst real distribution (red) actually looks much similar to the worst generation distribution. It may be more robust and meaningful to minimize  in the worst-case setting the departure of the real data distribution and the fake data distribution, which is conducted in our RGAN. Third, (c) shows that the real data distribution varies largely from the generated data points obtained by traditional GANs, indicating  the poor generalization of  the traditional GAN; in comparison, with a robust optimization in the worst-case setting, (d) demonstrates that the generated data look very close to the real data.




To clearly examine the visual quality, we demonstrate some images generated by WGAN-GP, DCGAN and their corresponding RGANs on  the CelebA dataset. These generated images are shown in complementary file for saving the space. General observations indicate that the existing GANs may sometimes lead to very bad generations. In comparison, with the robust training under the worst-case distribution, such very bad examples can hardly be seen in RGAN. This clearly demonstrates the advantages of the proposed model.

\section{Conclusion}
 In this paper, we consider the generalization issue of GANs and  propose a robust generative adversarial network model (RGAN).  We have designed a robust optimization framework where
  the generator and discriminator compete with each other in a worst-case setting.  The generator tries to map \textit{the worst input distribution} (rather than a specific input distribution) to real data distribution, while the discriminator attempts to distinguish the real and fake distribution \textit{with the worst perturbation}. We have provided theories showing that better generalization of the new robust framework can be guaranteed agaist traditional GANs. We also have conducted extensive experiments on CIFAR-10, STL-10 and CelebA datasets with two criteria (Inception score and FID) indicating that our proposed robust framework can improve consistently on several baseline GAN models.

\bibliographystyle{aaai}
\bibliography{reference1}

\section{\textbf{Appendix}}

\section{Proof for the optimal discriminator and global minimum of the utility function}
\textbf{Lemma 4.1}\ \
\textit{For an arbitrary fixed G, the optimal D of the game defined by the utility function
$V(G,D)$ is:}
\begin{eqnarray}
\begin{aligned}
D^*_G(x)=\frac{p_r^{\lambda}(x)}{p_r^{\lambda}(x)+p_g^{\lambda}(x)}
\label{eq:7a}
\end{aligned}
\end{eqnarray}
\textit{where $p_r^{\lambda}(x)=(1-\lambda)p_r+\lambda p'_r$ is the mixture distribution for real data with $\lambda\in [0,1]$. $p'_r$ is the worst distribution defined by $p'_r=\arg\min_{P:W(P,P_r)\leq \rho_r}\mathbb{E} _{x\sim P}[logD(x)]$. $p_g^{\lambda}(x)=(1-\lambda)p_g+\lambda p'_g$ is the mixture distribution for fake data. The worst distribution $p'_g$ is defined by $p'_g=\arg\min_{P:W(P,P_g)\leq \rho_g}\mathbb{E} _{G'\sim P}[1-logD(G'(z))]$.}
\label{lem_d1a}
\\\\
\textbf{Proof:}
Given the discriminator $D$ and generator $G$, the utility function can be rewritten as
\begin{eqnarray}
\begin{aligned}
V(G,D)&\triangleq (1-\lambda)[\mathbb{E} _{x\sim P_r}[logD(x)]\\&+\mathbb{E}_{G\sim P_g}[(1-logD(G(z_i)))]]\\&+\sup_{P:W(P,P_r)\leq \rho_r}\lambda\mathbb{E} _{x\sim P}[logD(x)]\\&+\sup_{P:W(P,P_g)\leq \rho_g}\lambda\mathbb{E}_{G'\sim P}[(1-logD(G'(z)))]
\\&=(1-\lambda)\int p_r(x)log(D(x))dx \\&+ (1-\lambda)\int p_g(x)log(1-D(x))dx \\&+ \lambda\int p'_r(x)log(D(x))dx \\&+ \lambda\int p'_g(x)log(1-D(x))dx
\\&= \int p^\lambda _r(x)log(D(x))dx \\&+ \int p^\lambda _g(x)log(1-D(x))dx
\label{eq:totala}
\end{aligned}
\end{eqnarray}
where $G'(z_i)=G(z_i)+r_2^i$ and $z_i\sim P_z$. $r_2^i$ is an arbitrary perturbation.
Then, it is easy to prove that the optimal $D$ is achieved at $D^*_G(x)=\frac{p_r^{\lambda}(x)}{p_r^{\lambda}(x)+p_g^{\lambda}(x)}$. \hfill $\blacksquare$
\newpage
\textbf{Lemma 4.2} \textit{When the optimum discriminator $D^*$ is achieved, the utility function reaches the global minimum if and only if $p_g^{\lambda}(x)=p_r^{\lambda}(x)$.}
\label{lem_d2a}
\\\\
\textbf{Proof:}
Given the optimal $D^*$, we can reformulate the function $V(G,D)$ as follows:
\begin{eqnarray}
\begin{aligned}
V(G,D^*) &= \int p^\lambda _r(x)log(\frac{p_r^{\lambda}(x)}{p_r^{\lambda}(x)+p_g^{\lambda}(x)})dx \\&+ \int p^\lambda _g(x)log(\frac{p_g^{\lambda}(x)}{p_r^{\lambda}(x)+p_g^{\lambda}(x)})dx
\\&= \int p^\lambda _r(x)log(\frac{p_r^{\lambda}(x)}{(p_r^{\lambda}(x)+p_g^{\lambda}(x))/2})dx \\&+ \int p^\lambda _g(x)log(\frac{p_g^{\lambda}(x)}{(p_r^{\lambda}(x)+p_g^{\lambda}(x))/2})dx -2log2
\\&= -2log2 + KL(p_r^{\lambda}(x)||(p_r^{\lambda}(x)+p_g^{\lambda}(x))/2) \\&+ KL(p_g^{\lambda}(x)||(p_r^{\lambda}(x)+p_g^{\lambda}(x))/2)
\label{eq:totala}
\end{aligned}
\end{eqnarray}

Then, $V(G,D^*)$ can be rewritten as:
\begin{eqnarray}
\begin{aligned}
V(G,D^*) = -2log2 + 2JSD(p^\lambda _r(x)||p^\lambda _g(x))
\label{eq:totala}
\end{aligned}
\end{eqnarray}
where $JSD$ is the Jensen-Shannon divergence. $JSD(p^\lambda _r(x)||p^\lambda _g(x))$ is always non-negative and its unique optimum
is achieved if and only if $p^\lambda _r(x)=p^\lambda _g(x)$.\hfill $\blacksquare$

\section{Proof for the generalizability upper bound of RGAN and original GAN}

In this section, we focus on proving that the generalizability upper bound of the discriminator in RGAN is tighter than that in traditional GANs.  We will omit the proof for the generator, since its proof is almost the same as that of the discriminator.

Before we prove the theory, we first introduce the objectives of the RGAN and  the original GAN as well as some definitions.

Formally, the training procedure of the original GAN can be formulated as:
\\ \\
\begin{eqnarray}
\begin{aligned}
\min_{G}\max_{D}S(G,D) &\triangleq \mathbb{E} _{x\sim P_r}[logD(x)]\\&+\mathbb{E}_{\widetilde{x}\sim P_g}[(1-logD(G(z)))]
\label{eq:1}
\end{aligned}
\end{eqnarray}
\\ \\
where $x$ and $\widetilde{x}=G(z)$ are real and fake examples sampled from the real data distribution $P_r$ and generation distribution $P_g$ respectively. The generation distribution is defined by $G(z)$ where $z\sim P_z$ ($P_z$ is a specific input noise distribution). The minmax problem cannot be solved directly since the expectation of the real and generation distribution is usually intractable. Therefore, an approximation problem can be defined as:
\\ \\
\begin{eqnarray}
\begin{aligned}
\min_{G}\max_{D} S_n(G,D) &\triangleq \frac{1}{n}\sum_{i=1}^{n}[logD(x_i)] \\&+\frac{1}{n}\sum_{i=1}^{n}[(1-logD(G(z_i)))]
\label{eq:2}
\end{aligned}
\end{eqnarray}
\\ \\
where $n$ examples of $x_i$ and $z_i$ are sampled from two specific distributions $P_r$ and $P_z$.\footnote{ Correction: the example number should be denoted as $n$ rather than $m$ as defined in the original submission.}  The mean value of the loss is used to approximate the original problem.

On the other hand, the objective of RGAN is formulated as:
\\ \\
 \begin{eqnarray}
 \footnotesize{
\begin{aligned}
\min_{G}\max_{D}&V(G,D)\triangleq (1-\lambda)S(G,D)\\ &+\sup_{P:W(P,P_r)\leq \rho_r}\lambda\mathbb{E} _{x\sim P}[logD(x)]\\&+\sup_{P:W(P,P_g)\leq \rho_g}\lambda\mathbb{E}_{G'\sim P}[(1-logD(G'(z)))]
\label{eq:total}
\end{aligned}}
\end{eqnarray}
\\ \\
where $G'(z_i)=G(z_i)+r_2^i$ and  $z_i\sim p^{\lambda}_z$. $p^{\lambda}_z$ is the mixture distribution defined by $p^{\lambda}_z=(1-\lambda)p_z+\lambda p'_z$ and $p'_z$ is the worst distribution defined by $p'_z=\arg\max_{P:W(P,P_z)\leq \rho_z}\mathbb{E} _{x\sim P}[1-logD(G(x))]$. $r_2^i$ is an arbitrary perturbation.
Again,the min-max problem of (\ref{eq:total}) is computationally intractable due to the expectations over real and fake distributions. An alternate way is to approximate the original problem with the empirical average of finite examples:

\begin{eqnarray}
 \footnotesize{
\begin{aligned}
\min_{G}\max_{D}V_n(G,D)&\triangleq (1-\lambda)S_n(G,D)\\ &+\frac{\lambda}{n}\sum_{i=1}^n [logD(x'_i)]\\ &+\frac{\lambda}{n}\sum_{i=1}^n[(1-logD(G'(z_i)))]
\label{eq:mean}
\end{aligned}}
\end{eqnarray}
\\ \\ \\
where $x'_i\sim p'_r$, $G'\sim p'_g$ and $z_i\sim p^{\lambda}_z$. $p^{\lambda}_z$ is the mixture distribution defined by $p^{\lambda}_z=(1-\lambda)p_z+\lambda p'_z$ and $p'_z$ is the worst distribution defined by $p'_z=\arg\max_{P:W(P,P_z)\leq \rho_z}\mathbb{E} _{x\sim P}[1-logD(G(x))]$.

\begin{defn}
Given a training set $S$ and data manifold $\mathcal{Z}$,  Algorithm $\mathcal{A}_s$ is $(K,\epsilon(S))$ robust, if  $\mathcal{Z}$ can be partitioned into $K$ disjoint sets, denoted as $\{C_i\}^K_{i=1}$, such that
$\forall$ $s\in S$,
\begin{eqnarray}
\footnotesize{
\begin{aligned}
s\in C_i, z\in C_i \longrightarrow |l(\mathcal{A}_s,s, y_s)-l(\mathcal{A}_s,z, y_z)|\leq \epsilon(S)
\label{eq:7}
\end{aligned}}
\end{eqnarray}
where $l(.)$ is the loss function~\cite{robust_and_generation}.
\label{assum}
\end{defn}

\begin{defn}
Algorithm $\mathcal{A}$ is $(\epsilon,S,\sigma)$ adversarial robust, if within the $\epsilon$-neighborhood of training set $S$, 
we have
\begin{eqnarray}
\footnotesize{
\begin{aligned}
\max_{s_a,s_b\in S, s'_b\in B(s_b,\epsilon)}|l(\mathcal{A},s_a,y_{s_a})-l(\mathcal{A},s'_b,y_{s_b})|\leq \sigma
\label{eq:7}
\end{aligned}}
\end{eqnarray}
where $y_{s_b}$ and $y_{s_a}$ are labels of training samples $s_a$ and $s_b$, and $s'_b$ is a perturbed sample in the region $B(s_b,\epsilon)$.
\label{assum}
\end{defn}

The notion of $(K,\epsilon(S))$ robust describes that the variation of the loss function can be bounded by $\epsilon(S)$ within each of $K$ disjoint sets on data manifold. The notion of $(\epsilon,S,\sigma)$ adversarial robust states that the difference between the loss of the natural example and its  perturbed counterpart can be bounded by $\sigma$ within the $\epsilon$-neighborhood of the training set.

The generalizability of the discriminator in our RGAN is defined to describe how fast the difference $|V^\theta_n-V^\theta|$ converges, where $V^\theta=\max_{D}V(G^*,D)$ and $V^\theta_n=\max_{D}V_n(G^*,D)$. Similarly, the generalizability of the discriminator in the original GAN can be defined by $|W^\theta_n-W^\theta|$, where $W^\theta=\max_{D}S(G^*,D)$ and $W^\theta_n=\max_{D}S_n(G^*,D)$.

\newpage
\textbf{Theorem 4.3}\footnote{Slightly different from the main text of our submission,  Theorem 4.3 and 4.4 have been minor revised here. However, the tighter bound conclusion of RGAN still stands. }
\textit{If the training set $S_d$ for discriminator consists of $n$ i.i.d samples, the discriminator of RGAN $D_r$ and the original GAN $D_{org}$ are both $(K,\epsilon(S_d))$ robust, $D_r$ is $(\epsilon,S_d,\sigma_{r})$ adversarial robust, and $D_{org}$ is $(\epsilon,S_d,\sigma_{org})$ adversarial robust, then, for any $\delta>0$ and  small enough $\epsilon$, with the probability at least $1-\delta$,  we have
\begin{eqnarray}
\begin{aligned}
&|V^\theta_n-V^\theta|\leq \gamma_1 \sigma_r + (1-\gamma_1)\epsilon(S_d) + M\sqrt{\frac{2Kln2+2ln(1/\delta)}{n}}
\nonumber\\
&|W^\theta_n-W^\theta|\leq \gamma_1 \sigma_{org} + (1-\gamma_1)\epsilon(S_d) + M\sqrt{\frac{2Kln2+2ln(1/\delta)}{n}}
\nonumber
\label{eq:72}
\end{aligned}
\end{eqnarray}
$D_r$ obtains the tighter upper bound than $D_{org}$, i.e.:
\begin{eqnarray}
\begin{aligned}
\gamma_1& \sigma_r + (1-\gamma_1)\epsilon(S_d) + M\sqrt{\frac{2Kln2+2ln(1/\delta)}{n}} \\ &\leq \gamma_1 \sigma_{org} + (1-\gamma_1)\epsilon(S_d) + M\sqrt{\frac{2Kln2+2ln(1/\delta)}{n}}
\nonumber
\label{eq:73}
\end{aligned}
\end{eqnarray}
where $\gamma_1\in [0,1]$, which is closely related to the intersection of $\epsilon$-neighborhood of training set $N_{\epsilon}=\bigcup_{s_i\in S_d}B(s_i,\epsilon)$ and data manifold $\mathcal{Z}$. $M$ is the upper bound of the loss of data manifold $\mathcal{Z}$.
}
\\\\
\textbf{Proof:}
Let $N_i$ be the set of index of points of training set $S_d=\{s_i\}_{i=1}^n$ that fall into the $C_i$ and $(|N_1|, ... , |N_K|)$ is an i.i.d multinomial random variable with parameters $n$ and $(\mu(C_1), ... , \mu(C_K))$.  The following holds by the Breteganolle-Huber-Carol inequality (cf Proposition A6.6 of~\cite{Van} ):
\begin{eqnarray}
\begin{aligned}
Pr\left\{\sum_{i=1}^{K}\left|\frac{N_i}{n}-\mu(C_i)\right|\geq \lambda \right\} \leq 2^K exp(\frac{-n\lambda^2}{2})
\label{eq:7}
\end{aligned}
\end{eqnarray}
Hence, with the probability at least $1-\delta$, we have:
\begin{eqnarray}
\footnotesize{
\begin{aligned}
\sum_{i=1}^{K}\left|\frac{N_i}{n}-\mu(C_i)\right|  \leq \sqrt{\frac{2Kln2+2ln(1/\delta)}{n}}
\label{eq:7}
\end{aligned}}
\end{eqnarray}
Since the discriminator of RGAN is $(K, \epsilon(S_d))$  robust and $(\epsilon, S_d, \sigma_r)$ adversarial  robust, the upper bound of  generalizability of the discriminator for RGAN
can be formulated as:
\begin{eqnarray}
\begin{aligned}
&\left|V^\theta_n-V^\theta\right| = |\sum_{i=1}^{K}\mathbb{E}(l(D_r,z,y_z)|z\in C_i)\mu(C_i)\\&-\frac{1}{n}\sum_{i=1}^{n}l(D_r,s_i,y_{s_i})|
\\ &\leq |\sum_{i=1}^{K}\mathbb{E}(l(D_r,z,y_z)|z\in C_i)\frac{\left|N_i\right|}{n}\\&-\frac{1}{n}\sum_{i=1}^{n}l(D_r,s_i,y_{s_i})|
\\ &+|\sum_{i=1}^{K}\mathbb{E}(l(D_r,z,y_z)|z\in C_i)\mu(C_i) \\&-\sum_{i=1}^{K}\mathbb{E}(l(D_r,z,y_z)|z\in C_i)\frac{\left|N_i\right|}{n}|
 \\ &= |\sum_{i=1}^{K}\lambda_i\mathbb{E}(l(D_r,z,y_z)|z\in C_i\cap \mathcal{N}_{\epsilon})\frac{\left|N_i\right|}{n}
\\ &+\sum_{i=1}^{K}(1-\lambda_i)\mathbb{E}(l(D_r,z,y_z)|z\in C_i\cap \mathcal{N}_{\epsilon}^{\mathrm{C}})\frac{\left|N_i\right|}{n}\\ &-\frac{1}{n}\sum_{i=1}^{K}\lambda_i\sum_{j\in N_i}l(D_r,s_j,y_{s_j})\\&-\frac{1}{n}\sum_{i=1}^{K}(1-\lambda_i)\sum_{j\in N_i}l(D_r,s_j,y_{s_j})|
\\ &+|\sum_{i=1}^{K}\mathbb{E}(l(D_r,z,y_z)|z\in C_i)\mu(C_i)
\\&-\sum_{i=1}^{K}\mathbb{E}(l(D_r,z,y_z)|z\in C_i)\frac{\left|N_i\right|}{n} |
\\ &\leq \left|\frac{1}{n}\sum_{i=1}^{K}\lambda_i\sum_{j\in N_i}\max_{z_1\in C_i\cap \mathcal{N}_{\epsilon}}\left|l(D_r,s_j,y_{s_j})-l(D_r,z_1,y_{z_1})\right|\right|
\\ &+\left|\frac{1}{n}\sum_{i=1}^{K}(1-\lambda_i)\sum_{j\in N_i}\max_{z_2\in C_i\cap \mathcal{N}_{\epsilon}^\mathrm{C}}\left|l(D_r,s_j,y_{s_j})-l(D_r,z_2,y_{z_2})\right|\right|
\\ &+\left|\max_{z\in Z}\left|l(D_r,z,y_z)\right|\sum_{i=1}^{K}\left|\frac{N_i}{n}-\mu(C_i)\right|\right|
\\ &\leq \left|\frac{1}{n}\sum_{i=1}^{K}\lambda_i\sum_{j\in N_i}\max_{s'\in \mathcal{N}_{\epsilon}}\left|l(D_r,s_j,y_{s_j})-l(D_r,s',y_{s'})\right|\right|
\\ &+\left|\frac{1}{n}\sum_{i=1}^{K}(1-\lambda_i)\sum_{j\in N_i}\max_{z_4\in C_i}\left|l(D_r,s_j,y_{s_j})-l(D_r,z_4,y_{z_4})\right|\right|
\\&+M\sum_{i=1}^{K}\left|\frac{N_i}{n}-\mu(C_i)\right|
\nonumber
\label{eq:72}
\end{aligned}
\end{eqnarray}
Here $\lambda_i$ represents the proportion of $C_i\cap \mathcal{N}_{\epsilon}$ in $C_i$. $\mathcal{N}_{\epsilon}^\mathrm{C}$ is the complementary set of $\mathcal{N}_{\epsilon}$ in the whole space and $\epsilon$ is small enough so that $B(s_i,\epsilon)$ does not intersect with data manifold of different classes. In other words, all samples in $B(s_i,\epsilon)$ share the same label $y_{s_i}$. Moreover, $D_r$ is $(K, \epsilon(S_d))$ robust and $(\epsilon, S_d, \sigma_r)$ adversarial robust. Then, we can further get the upper bound:
\begin{eqnarray}
\begin{aligned}
&|V^\theta_n-V^\theta|\leq \gamma_1 \sigma_r + (1-\gamma_1)\epsilon(S_d) + M\sqrt{\frac{2Kln2+2ln(1/\delta)}{n}}
\nonumber
\label{eq:72}
\end{aligned}
\end{eqnarray}
where $M$ is the upper bound of the loss on data Manifold $\mathcal{Z}$. $\gamma_1=\sum_{i=1}^{K}\lambda_i \frac{|N_i|}{n}$ and $\gamma_1\in [0,1]$.

Similarly, the upper bound for the generalizability of discriminator of original GANs can be formulated:
\begin{eqnarray}
\begin{aligned}
&|W^\theta_n-W^\theta|
\leq |\sum_{i=1}^{K}\mathbb{E}(l(D_{org},z,y_z)|z\in C_i)\frac{|N_i|}{n}\\&-\frac{1}{n}\sum_{i=1}^{n}l(D_{org},s_i,y_{s_i})|
\\ &+|\sum_{i=1}^{K}\mathbb{E}(l(D_{org},z,y_z)|z\in C_i)\mu(C_i)\\&-\sum_{i=1}^{K}\mathbb{E}(l(D_{org},z,y_z)|z\in C_i)\frac{|N_i|}{n}|
 \\ &= |\sum_{i=1}^{K}\lambda_i\mathbb{E}(l(D_{org},z,y_z)|z\in C_i\cap \mathcal{N}_{\epsilon})\frac{\left|N_i\right|}{n}
\\ &+\sum_{i=1}^{K}(1-\lambda_i)\mathbb{E}(l(D_{org},z,y_z)|z\in C_i\cap \mathcal{N}_{\epsilon}^{\mathrm{C}})\frac{\left|N_i\right|}{n}\\ &-\frac{1}{n}\sum_{i=1}^{K}\lambda_i\sum_{j\in N_i}l(D_{org},s_j,y_{s_j})-\frac{1}{n}\sum_{i=1}^{K}(1-\lambda_i)\sum_{j\in N_i}l(D_{org},s_j,y_{s_j})|
\\ &+|\sum_{i=1}^{K}\mathbb{E}(l(D_{org},z,y_z)|z\in C_i)\mu(C_i)
\\&-\sum_{i=1}^{K}\mathbb{E}(l(D_{org},z,y_z)|z\in C_i)\frac{\left|N_i\right|}{n} |
\\ &\leq \gamma_1 \sigma_{org} + (1-\gamma_1)\epsilon(S_d) + M\sqrt{\frac{2Kln2+2ln(1/\delta)}{n}}
\nonumber
\label{eq:7}
\end{aligned}
\end{eqnarray}
where $\gamma_1$ is defined the same as the above.

We can start to compare these two upper bounds. $D_r$ tries to minimize the mixture of losses for natural and perturbed samples: $\lambda l(D, s_i, y_{s_i})+ (1-\lambda)l(D, s'_i, y_{s_i})$ for $s_i\in S_d$, where $s'_i= \arg\min_{s\in B(s_i,\epsilon)} l(D, s, y_{s_i})$ is the worst perturbed samples within $B(s_i,\epsilon)$. $D_{org}$ just minimizes
the loss for natural example: $l(D, s_i, y_{s_i})$. Therefore, we have $\lambda l(D_r, s_i, y_{s_i})+ (1-\lambda)l(D_r, s'_i, y_{s_i})\leq \lambda l(D_{org}, s_i, y_{s_i}) + (1-\lambda)l(D_{org}, s'_i, y_{s_i})$ and $l(D_{org}, s_i, y_{s_i})\leq l(D_r, s_i, y_{s_i})$. Then, it is easy to get: $l(D_{r}, s'_i, y_{s_i})\leq l(D_{org}, s'_i, y_{s_i})$ for $s_i\in S_d$. Therefore, we have $\sigma_r\leq \sigma_{org}$.

Therefore, $D_r$ obtains the tighter upper bound:
\begin{eqnarray}
\begin{aligned}
\gamma_1& \sigma_r + (1-\gamma_1)\epsilon(S_d) + M\sqrt{\frac{2Kln2+2ln(1/\delta)}{n}} \\ &\leq \gamma_1 \sigma_{org} + (1-\gamma_1)\epsilon(S_d) + M\sqrt{\frac{2Kln2+2ln(1/\delta)}{n}}
\nonumber
\label{eq:73}
\end{aligned}
\end{eqnarray}\hfill $\blacksquare$
\\
\\

Now we start to analyze the effect of $\epsilon$ on the generalizability upper bound of  RGAN. When $\epsilon$ is large so that $B(s_i, \epsilon)$ contains samples with the  label different from  $y_{s_i}$, $D_r$ tries to minimize $\lambda l(D, s_i, y_{s_i})+ (1-\lambda)l(D, s'_i, y_{s_i})$ which enforces the prediction of $s'_i$ to the wrong label $y_{s_i}$. In other words, the samples with the different label are predicted to the wrong label $y_{s_i}$. When $\epsilon$ is small enough that all samples within $B(s_i,\epsilon)$ share the same label $y_{s_i}$, the generalizability of intersection between $N_{\epsilon}$ and $\mathcal{Z}$ can be bounded by $\sigma_r$. In other words, the better generalizability bound can be guaranteed by a smaller $\sigma_r$.

Similarly, the generalizability of the generator for RGAN can be defined as the convergence of  $|Q^\phi_n-Q^\phi|$, where $Q^\phi=\min_{G}V(G,D^*)$ and $Q^\phi_n=\min_{G}V_n(G,D^*)$. For original GANs, the generalizability of the generator is defined by $|U^\phi_n-U^\phi|$, where $U^\phi=\min_{G}S(G,D^*)$ and $U^\phi_n=\min_{G}S_n(G,D^*)$.  We can also prove that the RGAN's generator has a tighter upper bound of the generalizability  than that of traditional GANs in the same way. We formulate the theory about the generator as follows but would omit its proof process.
\\

\textbf{Theorem 4.4}\textit{
If the training set $S_g$ for the generator consists of $n$ i.i.d samples, the generator of our proposed method $G_r$ and original GAN $G_{org}$ are $(K,\epsilon(S_g))$ robust, $G_r$ is $(\epsilon',S_g,\sigma'_{r})$ adversarial robust and $G_{org}$ is $(\epsilon',S_g,\sigma'_{org})$ adversarial robust, then, for any $\delta>0$ and small enough $\epsilon'$, with the probability at least $1-\delta$,  we have
\begin{eqnarray*}
\begin{aligned}
|Q^\phi_n-Q^\phi|\leq \gamma'_1 \sigma'_r + (1-\gamma'_1)\epsilon(S_g) + M'\sqrt{\frac{2Kln2+2ln\frac{1}{\delta}}{n}}\\
|U^\phi_n-U^\phi|\leq \gamma'_1 \sigma'_{org} + (1-\gamma'_1)\epsilon(S_g) + M'\sqrt{\frac{2Kln2+2ln\frac{1}{\delta}}{n}}
\end{aligned}
\end{eqnarray*}
$G_r$ obtains the tighter upper bound than $G_{org}$, i.e.:
\begin{eqnarray*}
&\gamma'_1 \sigma'_r + (1-\gamma'_1)\epsilon(S_g) + M'\sqrt{\frac{2Kln2+2ln\frac{1}{\delta}}{n}}\\
&\leq \gamma'_1 \sigma'_{org} + (1-\gamma'_1)\epsilon(S_g) + M'\sqrt{\frac{2Kln2+2ln\frac{1}{\delta}}{n}}
\end{eqnarray*}
where $\gamma'_1\in [0,1]$, which are closely related to the intersection of $\epsilon'$-neighborhood of the training set $N_{\epsilon'}=\bigcup_{s_i\in S_g}B(s_i,\epsilon')$ and data manifold $\mathcal{Z}'$. $M'$ is the upper bound of the loss of data manifold $\mathcal{Z}'$.
}

\begin{figure*}[ht]
\centering
\subfigure[WGAN-GP] {
\includegraphics[width=0.45\textwidth]{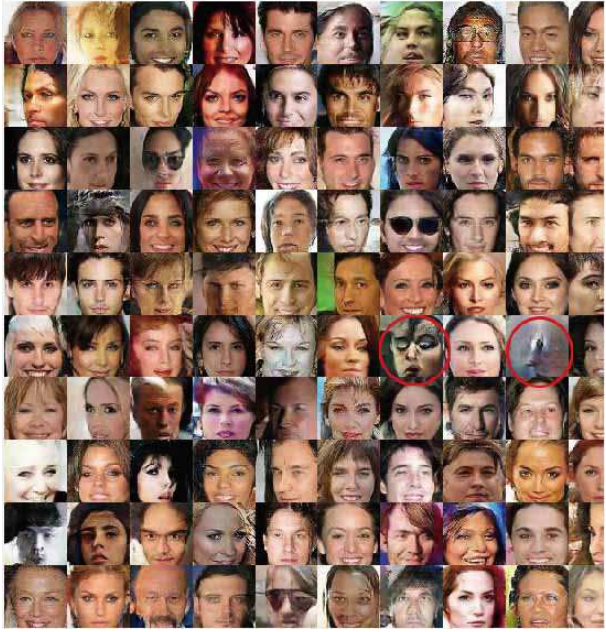}
\label{fig:ad1} }
\subfigure[RGAN (WGAN-GP)] {
\includegraphics[width=0.45\textwidth]{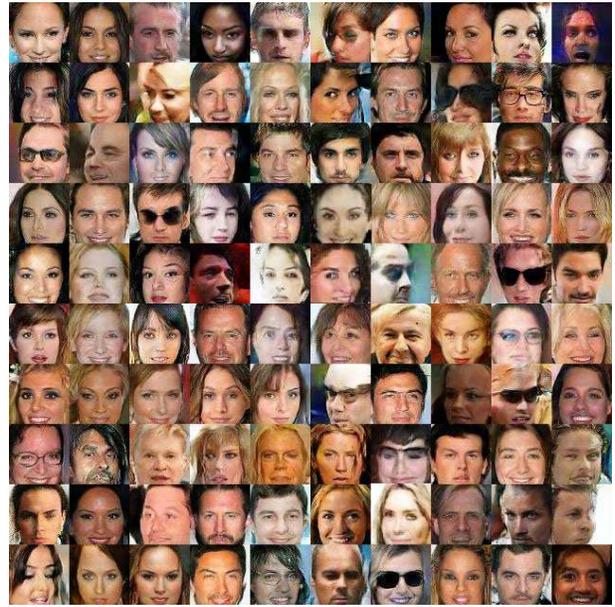}
\label{fig:ad2} }
\subfigure[DCGAN.] {
\includegraphics[width=0.45\textwidth]{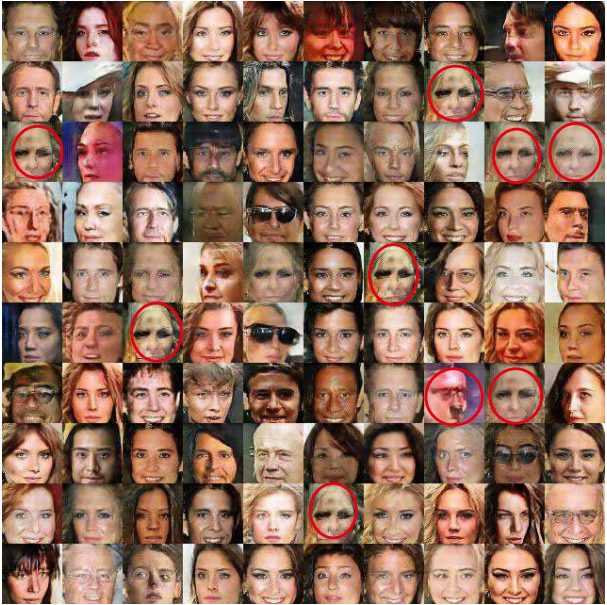}
\label{fig:ad3} }
\subfigure[RGAN (DCGAN)] {
\includegraphics[width=0.45\textwidth]{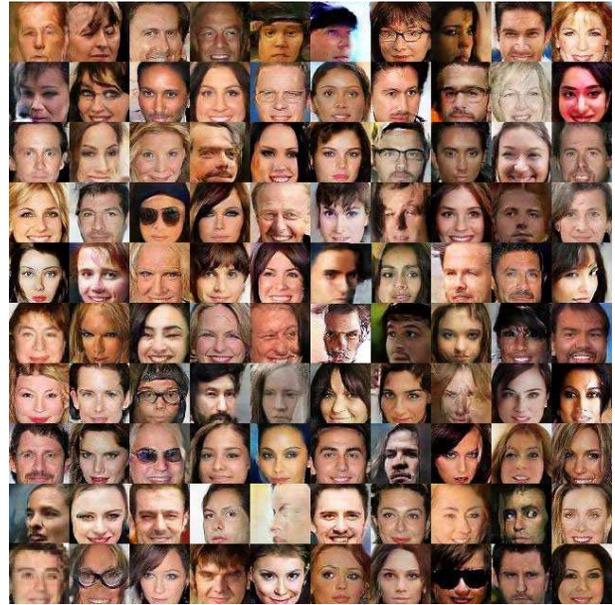}
\label{fig:ad3} }

\caption{Face images generated by WGAN-GP, DCGAN and the corresponding RGANs. In (a), WGAN-GP generates two obviously strange faces highlighted with red circles. In (c), several repeated low quality faces are generated by DCGAN highlighted by red circles. Our method achieves visually better and more stable results.}
\label{fig_AE}
\label{tsne_Face}
\end{figure*}

\section{Visualization on CelebA}

We have shown quantitative comparisons between our proposed model and the other competitive methods in our submission. We now show some qualitative results to  examine the different models visually. Particularly, we demonstrate some images generated by WGAN-GP, DCGAN and their corresponding RGANs on the CelebA dataset. These generated images are shown in Figure~\ref{tsne_Face}. As we can observe from these examples, the existing GANs may sometimes lead to very bad generations as circled in (a) and (c). In comparison, with the robust training under the worst-case distribution, such very bad examples can hardly be seen in RGAN. This demonstrates the robustness of the proposed model.


\end{document}